\documentclass[pdflatex,sn-mathphys-num]{sn-jnl}

\usepackage{graphicx}%
\usepackage{multirow}%
\usepackage{amsmath,amssymb,amsfonts}%
\usepackage{amsthm}%
\usepackage{mathrsfs}%
\usepackage[title]{appendix}%
\usepackage{xcolor}%
\usepackage{textcomp}%
\usepackage{booktabs}%
\usepackage{algorithm}%
\usepackage{algorithmicx}%
\usepackage{algpseudocode}%
\usepackage{listings}%
\usepackage[utf8]{inputenc}
\usepackage[T1]{fontenc}
\usepackage{lmodern}
\usepackage{url}
\usepackage{mathtools}
\usepackage{nicefrac}
\usepackage{microtype}
\usepackage{longtable}
\usepackage{array}
\usepackage{hyperref}

\theoremstyle{thmstyleone}%
\newtheorem{theorem}{Theorem}
\newtheorem{proposition}[theorem]{Proposition}%

\theoremstyle{thmstyletwo}%
\newtheorem{remark}{Remark}%

\theoremstyle{thmstylethree}%
\newtheorem{definition}{Definition}%

\begin{document}

\title[The Multiclass Score-Oriented Loss (MultiSOL)]{The Multiclass Score-Oriented Loss (MultiSOL) \\ on the Simplex}


\author*[1]{\fnm{Francesco} \sur{Marchetti}}\email{francesco.marchetti@unipd.it}
\author[2]{\fnm{Edoardo} \sur{Legnaro}}\email{edoardo.legnaro@edu.unige.it}
\author[2]{\fnm{Sabrina} \sur{Guastavino}}\email{sabrina.guastavino@unige.it}

\affil*[1]{\orgdiv{Department of Mathematics ``Tullio Levi-Civita""}, \orgname{University of Padova}, \orgaddress{\city{Padova}, \country{Italy}}}
\affil[2]{\orgdiv{Department of Mathematics DIMA}, \orgname{University of Genova}, \orgaddress{\city{Genova}, \country{Italy}}}


\abstract{In the supervised binary classification setting, score-oriented losses have been introduced with the aim of optimizing a chosen performance metric directly during the training phase, thus avoiding \textit{a posteriori} threshold tuning. To do this, in their construction, the decision threshold is treated as a random variable provided with a certain \textit{a priori} distribution. In this paper, we use a recently introduced multidimensional threshold-based classification framework to extend such score-oriented losses to multiclass classification, defining the Multiclass Score-Oriented Loss (MultiSOL) functions. As also demonstrated by several classification experiments, this proposed family of losses is designed to preserve the main advantages observed in the binary setting, such as the direct optimization of the target metric and the robustness to class imbalance, achieving performance comparable to other state-of-the-art loss functions and providing new insights into the interaction between simplex geometry and score-oriented learning.}




\maketitle

\section{Introduction}

In supervised classification problems addressed by neural networks, the choice of the loss function plays a central role, as it guides the optimization of the network during the training phase. In binary classification, after the network has been trained, tuning the classification threshold in the output range $(0,1)$ to optimize the performance of the network in terms of a chosen score of interest represents a well-established practice, as it can severely improve the measured score especially in particular cases, such as, class-imbalanced settings. Alternatively, Score-Oriented Losses (SOLs) have been introduced in \cite{Marchetti22} to provide automatic optimization of a score in the training phase, without the need for a posteriori maximization. This approach, which is based on the treatment of the threshold as a random variable endowed with a chosen \textit{a priori} probability density function, has been further extended to weighted scores and the multilabel framework \cite{marchetti2025comprehensive}, and applied in fields where class unbalancing is a typical issue \cite{guastavino2024forecasting,guastavino2023operational,guastavino2022implementation,vong2025bypassing}. Related work includes, for example, \cite{han2024anyloss}, where the introduced and extensively analyzed \textit{Anyloss} can be included in the SOL framework by choosing the \textit{a priori} logistic distribution.
When trying to extend the SOL approach to the multiclass scenario, the different nature of the consolidated \textit{argmax} classification rule prevents the construction of a straightforward generalization. Indeed, unlike the binary case where the threshold-based classification rule is applied on top of a sigmoid output unit, when dealing with multiple classes the \textit{argmax} rule used on softmax output units is threshold-independent as it looks at the class with the highest predicted value \cite{murphy2018machine}, thus depriving a candidate of extended score-oriented loss of its key ingredient, that is, the probabilistic interpretation of the decision threshold. Other multiclass classification rules have been proposed \cite{bridle1990probabilistic,martins2016softmax}, such as the Fréchet mean \cite{roberts2023geometry} or the \lq\lq inflated argmax\rq\rq \cite{soloff2024building}, but still lack some multidimensional decision threshold, which also limits the opportunity of some \textit{a posteriori} fine-tuning of a score.

To overcome this issue, in \cite{Legnaro25Multiclass} a different perspective has been proposed for the interpretation of the output of the model. More specifically, instead of the probabilistic interpretation underpinning the application of the \textit{argmax} rule, the authors interpreted the softmax output in its natural domain, i.e., the multidimensional simplex, and designed a classification rule that is governed by a multidimensional threshold parameter.

In this work, we build on this multiclass threshold-based framework to introduce the Multi-class Score-Oriented Loss function (MultiSOL) as the extension of the SOL to the multiclass setting. Just as in the binary case, MultiSOL enables direct optimization of a desired skill score during training by treating the threshold as a multivariate random variable. In our analysis with various Dirichlet distributions on the simplex considered, we show how the choice of the \textit{a priori} distribution for the threshold drives the optimization dynamics in the training phase, as certified by the carried out \textit{a posteriori} analysis. In addition to evaluating the MultiSOL properties in terms of the original binary SOL ones, we also compare MultiSOL with the classical Categorical Cross-Entropy (CE) and with other methods investigated in the wide research line concerning loss function design. Among them are the $\alpha$-divergence loss with GAN regularization \cite{pmlr-v235-novello24a}, the Label-Distribution-Aware Margin loss \cite{cao2019learning}, and the Influence-Balanced loss \cite{park2021influence}, to cite a few. These losses aim to improve robustness to class imbalance, calibration, and generalization, but none of them directly optimizes a performance score of interest. From our experiments, MultiSOL results to be competitive with state-of-the-art loss functions, and also offers a principled mechanism for score-oriented optimization in the multiclass context, which is particularly effective and useful in applications with imbalanced classes \cite{pathmnist,tissuemnist,legnaro2025deep}. 

The paper is organized as follows.
Section \ref{sec:binary} recalls the binary score-oriented loss framework and the key ideas that motivate its extension.
Section \ref{sec:multi_sol} introduces the multiclass threshold-based formulation on the simplex and develops the corresponding MultiSOL functions, together with their analytical properties and computational construction.
Section \ref{sec:results} presents an extensive set of experiments assessing MultiSOL behaviour, the influence of the chosen prior distribution, its score-oriented effects, and its performance relative to state-of-the-art loss functions.
The results are interpreted and discussed in Section \ref{sec:discussion}.
Final remarks and perspectives for future research are reported in Section \ref{sec:conclusions}.
Additional dataset information is provided in Appendix~\ref{app:datasets}.

\section{Score-oriented losses in the binary case}\label{sec:binary}

In the binary classification setting, each data sample $\boldsymbol{x}_i$ belonging to the $d$-dimensional dataset 
$\mathcal{X}=\{\boldsymbol{x}_1,\dots,\boldsymbol{x}_n\}\subset\Omega\subset\mathbb{R}^d$ is associated with a label 
$y_i\in\{0,1\}$. The aim is to learn the underlying mapping between samples and labels through a neural 
network that outputs
\begin{equation}\label{eq:output}
   \hat{y}_{\boldsymbol{\theta}}(\boldsymbol{x}) = (\sigma \circ h)(\boldsymbol{x},\boldsymbol{\theta}) \in [0,1], \quad \boldsymbol{x}\in\Omega,
\end{equation}
where $h(\boldsymbol{x},\boldsymbol{\theta})$ denotes the transformation computed by the \textit{input} and \textit{hidden} layers, parameterized 
by the weight vector (or matrix) $\boldsymbol{\theta}$. The function $\sigma$ represents the standard 
\textit{sigmoid} activation, defined as
$\sigma(h)=(1+e^{-h})^{-1}$.

During the training process, the weights of the network are tuned in such a way that a certain objective function is minimized, that is, we consider the problem
\begin{equation}\label{eq:obj_fun}
    \min_{\boldsymbol{\theta}}\ell(\hat{y}_{\boldsymbol{\theta}}(\boldsymbol{x}),y)+\gamma R(\boldsymbol{\theta}),\; \boldsymbol{x}\in\Omega,
\end{equation}
where $\ell$ is a chosen loss function and $R(\boldsymbol{\theta})$ is a possible regularization term that is controlled by a parameter $\gamma\in\mathbb{R}_{>0}$ called regularization parameter \cite{girosi1995regularization}. After that, an unseen test sample $\boldsymbol{x}$ is classified by assigning its corresponding output $\hat{y}_{\boldsymbol{\theta}}(\boldsymbol{x})$ to one of the two labels $\{y=0\}$ or $\{y=1\}$ by passing $\hat{y}=\hat{y}_{\boldsymbol{\theta}}(\boldsymbol{x})$ through a function
\begin{equation*}
    \mathbf{1}_{\hat{y}}(\tau)=\mathbf{1}_{\{\hat{y}>\tau\}}=\begin{dcases} 0 & \textrm{if } \hat{y}\le \tau,\\
        1 & \textrm{if }\hat{y}>\tau,\end{dcases}
\end{equation*}
being $\tau\in(0,1)$ a threshold value.

Especially in unbalanced settings, carefully tuning the threshold value is crucial in order to obtain a classifier that is well-performing when measured in terms of a score of interest, whose choice may depend on the setting. We recall the following.
\begin{definition}\label{def:score}
A score $s$ is a function that takes in input the entries of the confusion matrix 
\begin{equation}\label{eq:cm}
\mathrm{CM}(\tau,\boldsymbol{\theta}) = 
\begin{pmatrix}
\mathrm{TN}(\tau,\boldsymbol{\theta}) & \mathrm{FP}(\tau,\boldsymbol{\theta}) \\
\mathrm{FN}(\tau,\boldsymbol{\theta}) & \mathrm{TP}(\tau,\boldsymbol{\theta})
\end{pmatrix},
\end{equation}
which is non-decreasing with respect to $\mathrm{TN}$ and $\mathrm{TP}$ and non-increasing with respect to $\mathrm{FN}$ and $\mathrm{FP}$, being
\begin{equation*}\label{eq:mat_defi}
    \begin{split}
         & \mathrm{TN}(\tau,\boldsymbol{\theta}) = \sum_{i=1}^n{(1-y_i)\mathbf{1}_{\{\hat{y}_{\boldsymbol{\theta}}(\boldsymbol{x}_i)<\tau\}}}, \quad \mathrm{TP}(\tau,\boldsymbol{\theta}) = \sum_{i=1}^n{y_i\mathbf{1}_{\{\hat{y}_{\boldsymbol{\theta}}(\boldsymbol{x}_i)>\tau\}}},\\
         & \mathrm{FP}(\tau,\boldsymbol{\theta}) = \sum_{i=1}^n{(1-y_i)\mathbf{1}_{\{\hat{y}_{\boldsymbol{\theta}}(\boldsymbol{x}_i)>\tau\}}},\quad \mathrm{FN}(\tau,\boldsymbol{\theta}) = \sum_{i=1}^n{y_i\mathbf{1}_{\{\hat{y}_{\boldsymbol{\theta}}(\boldsymbol{x}_i)<\tau\}}}.
        \end{split}
\end{equation*}
\end{definition}
When carrying out a posteriori maximization of a score $s=s(\mathrm{CM}(\tau,\boldsymbol{\theta}))$, letting $\boldsymbol{\theta}^{\star}$ be the solution found for \eqref{eq:obj_fun}, after the training procedure we compute
\begin{equation*}\label{eq:a_posteriori}
    \max_{\tau\in(0,1)}s(\mathrm{CM}(\tau,\boldsymbol{\theta}^\star)).
\end{equation*}
Such a tuning, which is carried out using the training or validation set, is very often important to significantly improve the classification score achieved then on the test set, and it represents a common practice in this framework.

Alternatively to threshold tuning, in \cite{Marchetti22} the authors defined what they called Score-Oriented Loss (SOL) functions. To construct them, the first crucial step is to let $\tau$ be a continuous random variable whose probability density function (pdf) $f$ is supported in $[a,b]\subseteq[0,1]$, $a,b\in\mathbb{R}$. We denote as $F$ the cumulative density function (cdf)
\begin{equation*}
    F(x)=\int_{a}^{x}{f(\xi)\mathrm{d}\xi},\quad x\le b.
\end{equation*}
This ensures the possibility of averaging the entries of  $\mathrm{CM}$ with respect to the threshold, thus replacing the \textit{irregular} indicator function with the \textit{regular} cdf $F$. Indeed, we recall that the expected value of the indicator function behaves as
\begin{equation*}
    \mathbb{E}_{\tau\sim f}[\mathbf{1}_{\{x>\tau\}}]=\int_{a}^{b}{\mathbf{1}_{\{x>\xi\}}f(\xi)\mathrm{d}\xi}=\int_{a}^{x}{f(\xi)\mathrm{d}\xi}=F(x).
\end{equation*}
As a consequence, we can consider an \textit{expected} confusion matrix
\begin{equation}\label{eq:matrices_sol}
    \mathbb{E}_{\tau\sim f}[\mathrm{CM}(\tau,\boldsymbol{\theta})]= \begin{pmatrix}
\mathbb{E}_{\tau\sim f}[\mathrm{TN}(\tau,\boldsymbol{\theta})] & \mathbb{E}_{\tau\sim f}[\mathrm{FP}(\tau,\boldsymbol{\theta})] \\
\mathbb{E}_{\tau\sim f}[\mathrm{FN}(\tau,\boldsymbol{\theta})] & \mathbb{E}_{\tau\sim f}[\mathrm{TP}(\tau,\boldsymbol{\theta})]
\end{pmatrix},
\end{equation}
where
\begin{equation*}
        \begin{split}
         & \mathbb{E}_{\tau\sim f}[\mathrm{TN}(\tau,\boldsymbol{\theta})] = \sum_{i=1}^n{(1-y_i)(1-F(\hat{y}_{\boldsymbol{\theta}}(\boldsymbol{x}_i)))}, \quad \mathbb{E}_{\tau\sim f}[\mathrm{TP}(\tau,\boldsymbol{\theta})] = \sum_{i=1}^n{y_iF(\hat{y}_{\boldsymbol{\theta}}(\boldsymbol{x}_i))},\\
         & \mathbb{E}_{\tau\sim f}[\mathrm{FP}(\tau,\boldsymbol{\theta})] = \sum_{i=1}^n{(1-y_i)F(\hat{y}_{\boldsymbol{\theta}}(\boldsymbol{x}_i))}, \quad \mathbb{E}_{\tau\sim f}[\mathrm{FN}(\tau,\boldsymbol{\theta})] = \sum_{i=1}^n{y_i(1-F(\hat{y}_{\boldsymbol{\theta}}(\boldsymbol{x}_i)))}.
        \end{split}
\end{equation*}
Then, we can define the following family of loss functions.
\begin{definition}
    Let $s$ be a classification score and $f$ the pdf chosen for $\tau$. We define a SOL function related to the score $s$, with a priori pdf $f$, the loss
\begin{equation*}
    \ell_{s}(\{\hat{{y}}_{\boldsymbol{\theta}}(\boldsymbol{x}_i)\}_{i:n},\{{y}_i\}_{i:n})= - s\big(\mathbb{E}_{\tau\sim f}[\mathrm{CM}(\tau,\boldsymbol{\theta})]\big),
\end{equation*}
where $\{\hat{{y}}_{\boldsymbol{\theta}}(\boldsymbol{x}_i)\}_{i:n}=\{\hat{{y}}_{\boldsymbol{\theta}}(\boldsymbol{x}_1),\dots,\hat{{y}}_{\boldsymbol{\theta}}(\boldsymbol{x}_n)\}$ and $\{{y}_i\}_{i:n}=\{{y}_1,\dots,{y}_n\}$ are the batch of predictions and the corresponding labels, respectively.
\end{definition}
Note that writing $ \ell_{s}(\{\hat{{y}}_{\boldsymbol{\theta}}(\boldsymbol{x}_i)\}_{i:n},\{{y}_i\}_{i:n})$ represents a slight abuse of notation (cf. \eqref{eq:obj_fun}), and it is introduced to deal with the \textit{batchwise} definition of SOLs without making the notation heavier.

When considering a score-oriented loss, we have (see \cite[Theorem 1]{marchetti2025comprehensive})
\begin{equation}\label{eq:resultone}
\min_{\boldsymbol{\theta}}\ell_{s}(\{\hat{{y}}_{\boldsymbol{\theta}}(\boldsymbol{x}_i)\}_{i:n},\{{y}_i\}_{i:n})\approx \max_{\boldsymbol{\theta}} \mathbb{E}_{\tau\sim f}[s(\mathrm{CM}(\tau,\boldsymbol{\theta}))].
\end{equation}
In previous work, two main intertwined properties of score-oriented losses were thus verified for the 
binary and multilabel frameworks.
\begin{enumerate}
	\item 
	\textit{Influence of a priori chosen pdf.} The pdf chosen a priori directly influences the expected value in the matrices \eqref{eq:matrices_sol}. Therefore, threshold values outside the support of the pdf are not taken into consideration, and the optimal threshold is very likely to be placed in the concentration areas of the a priori distribution.
	\item
	\textit{Automatic a posteriori maximization of the score.} The effect provided by the a priori pdf leads to an \textit{automatic} a posteriori optimization of the chosen score directly in the training phase, in the \textit{averaged} sense provided by \eqref{eq:resultone}.
\end{enumerate}

In the next section, our purpose is to extend the score-oriented approach to the multiclass framework.

\section{Multiclass score-oriented losses}\label{sec:multi_sol}

\subsection{Threshold-based multiclass framework}\label{sec:multi_simplex}

Before introducing a key ingredient for the construction of multiclass score-oriented losses, that is, the threshold-based framework in the multidimensional simplex, we briefly recall the classical multiclass scenario, and discuss it in relationship with SOLs.

In a standard multiclass problem, where $C_1,\dots,C_m$ are $m>2$ classes and each $y_i\in C_j$ for a unique $j=1,\dots,m$, the classification is no longer based on a threshold value. Instead, $\sigma$ (see \eqref{eq:output}) is a \textit{softmax} activation function, which models the output as 
\begin{equation*}
    \hat{\boldsymbol{y}}_i = (\hat{y}_i^1,\dots,\hat{y}_i^m),\quad \sum_{j=1}^m \hat{y}_i^j = 1,
\end{equation*}
and the sample $\boldsymbol{x}_i$ is assigned to the class $j^\star$ that represents the \textit{argmax} of $\hat{\boldsymbol{y}}_i$ with respect to the $m$ classes, that is, the output is interpreted as a probability distribution and $j^\star = \textrm{argmax}_{j=1,\dots,m}\hat{y}_i^j$. 

In order to extend SOLs to deal with multiclass tasks, the argmax-based classification rule is not suitable, mainly because it does not build on a threshold. This is a known issue for other analysis tools such as, e.g., Receiver Operating Characteristic (ROC) curves, and the most widely used solution to extend a native binary method to multiple classes consists of splitting the multiclass problem in $m$ \textit{one-vs-rest} binary tasks. We remark that this is the case of binary scores, which are computed in the multiclass case by considering $m$ \textit{one-vs-rest} confusion matrices \eqref{eq:cm}, where a class is considered positive and the set of remaining ones as negative. However, introducing then multiple independent thresholds for each derived binary problem would be \textit{unnatural}, as the intertwined nature of these problems would be lost, thus leading to a sort of mixed multiclass-multilabel approach.

To overcome this issue, in \cite{Legnaro25Multiclass} a threshold-based framework has been introduced for the multiclass scenario and will be outlined in the following lines. First, we observe that the output $\hat{\boldsymbol{y}}$ is contained in the $(m-1)$-simplex $S_m=\{\boldsymbol{z}\in\mathbb{R}^m\:|\: \sum_{j=1}^m \boldsymbol{z}^j = 1 \}$, whose vertices are the \textit{one-hot} encoded classes $\boldsymbol{e}_1,\dots,\boldsymbol{e}_m$ corresponding to $C_1,\dots,C_m$ \cite{Harris13}. Then, as in the binary case the threshold value establishes which subinterval of $(0,1)$ is assigned to the zero class, automatically assigning the remaining region to the other class, the next definition indicates which areas of $S_m$ are assigned to each considered class (we refer to \cite[Definitions 1 \& 2]{Legnaro25Multiclass}).

\begin{definition}\label{def:region}
Let $\boldsymbol{\tau}=(\tau^1,\dots,\tau^m)\in S_m$ be a multidimensional threshold. A simplex classification collection $R_1(\boldsymbol{\tau}),\dots,R_m(\boldsymbol{\tau})$ of classification regions $R_j(\boldsymbol{\tau})$ for the class $C_j$, $j=1,\dots,m$, satisfies the following properties.
\begin{enumerate}
    \item
     $R_j(\boldsymbol{\tau})\subset S_m$ for $j=1,\dots,m$,
     \item 
     $\boldsymbol{e}_j\in R_j(\boldsymbol{\tau})$ for $j=1,\dots,m$,
     \item
     $\textrm{cl}\big(\bigcup_{j=1}^m{R_j(\boldsymbol{\tau})}\big)=S_m$,
\end{enumerate}
where $cl$ denotes the topological closure.
Moreover, it is called a proper simplex classification collection if 
\begin{equation*}
     R_k(\boldsymbol{\tau})\cap R_j(\boldsymbol{\tau})=\emptyset 
\end{equation*}
for $k,j=1,\dots,m$, $k\neq j$.
\end{definition}
According to the definition above, a proper classification collection consists of a sort of partition of the simplex $S_m$, where each output is assigned to a unique class, besides null measure subsets (this ambiguity applies to the argmax rule too with, e.g., the output $(0.4,0.4,0.2)$). This provides well posedness to the approach, and varying $\boldsymbol{\tau}$ will produce a variation in the corresponding \textit{classification rule}, analogously to the binary setting.

\begin{figure}[htbp]
    \centering
    \includegraphics[width=0.7\linewidth]{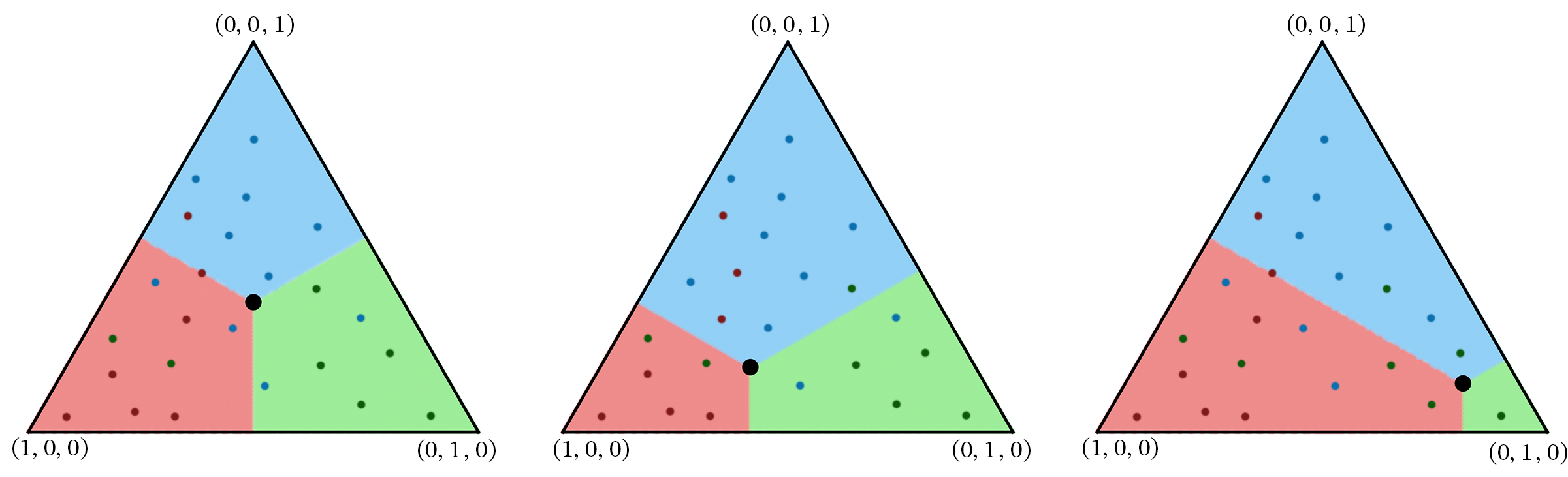}
    \caption{For $m=3$, the three regions $R_1(\boldsymbol{\tau})$ (red), $R_2(\boldsymbol{\tau})$ (green) and $R_3(\boldsymbol{\tau})$ (blue). From left to right: $\boldsymbol{\tau}=(1/3,1/3,1/3)$ (i.e., the argmax rule), $\boldsymbol{\tau}=(1/2,1/3,1/6)$, $\boldsymbol{\tau}=(1/8,3/4,1/8)$ (big black dot). The color blue, red or green represents the true label of the samples (colored dots). Number of misclassifications from left to right: $7$, $8$ and $10$ (from \cite{Legnaro25Multiclass}).}
    \label{fig:simplex_points}
\end{figure}

Among the possible proper classification collections that can be defined in the simplex, we then consider
\begin{equation}\label{eq:natural}
    R_j(\boldsymbol{\tau})=\{\boldsymbol{z}\in S_m\:|\: z^j-z^k>\tau^j-\tau^k,\; k\neq j\},\; j=1,\dots,m.
\end{equation}
We will refer to this choice as the \textit{natural} classification collection, since it is proper and directly generalizes the argmax procedure (see \cite[Proposition 1]{Legnaro25Multiclass}). In Figure \ref{fig:simplex_points}, we depict some examples of classification collections with the form of \eqref{eq:natural}.

\subsection{Designing the MultiSOL}

In the following, in analogy to binary SOLs, our aim is to take advantage of the threshold-based approach introduced in the previous subsection for multiclass problems in order to define multiclass SOLs (MultiSOLs).

Let $\boldsymbol{\tau}$ be a continuous random variable whose multivariate probability density function (pdf) $f$ is supported in $S_m$, consider for each $j=1,\dots,m$ the expected matrix
\begin{equation*}
	\mathbb{E}_{\boldsymbol{\tau}\sim f}[\mathrm{CM}_j(\boldsymbol{\tau},\boldsymbol{\theta})] = 
	\begin{pmatrix}
			\mathbb{E}_{\boldsymbol{\tau}\sim f}[\mathrm{TN}_j(\boldsymbol{\tau},\boldsymbol{\theta})] & 	\mathbb{E}_{\boldsymbol{\tau}\sim f}[\mathrm{FP}_j(\boldsymbol{\tau},\boldsymbol{\theta})] \\
			\mathbb{E}_{\boldsymbol{\tau}\sim f}[\mathrm{FN}_j(\boldsymbol{\tau},\boldsymbol{\theta})] & 	\mathbb{E}_{\boldsymbol{\tau}\sim f}[\mathrm{TP}_j(\boldsymbol{\tau},\boldsymbol{\theta})]
	\end{pmatrix},
\end{equation*}
with
\begin{equation*}
	\begin{split}
		& 	\mathbb{E}_{\boldsymbol{\tau}\sim f}[\mathrm{TN}_j(\boldsymbol{\tau},\boldsymbol{\theta})] = \sum_{i=1}^n{\mathbf{1}_{\{\boldsymbol{y}_i\neq \boldsymbol{e}_j\}}	\mathbb{E}_{\boldsymbol{\tau}\sim f}[\mathbf{1}_{\{\hat{\boldsymbol{y}}_{\boldsymbol{\theta}}(\boldsymbol{x}_i)\notin R_j(\boldsymbol{\tau})\}}]},\\ 
		& 	\mathbb{E}_{\boldsymbol{\tau}\sim f}[\mathrm{TP}_j(\boldsymbol{\tau},\boldsymbol{\theta})] = \sum_{i=1}^n{\mathbf{1}_{\{\boldsymbol{y}_i= \boldsymbol{e}_j\}}	\mathbb{E}_{\boldsymbol{\tau}\sim f}[\mathbf{1}_{\{\hat{\boldsymbol{y}}_{\boldsymbol{\theta}}(\boldsymbol{x}_i)\in R_j(\boldsymbol{\tau})\}}]},\\
		& 	\mathbb{E}_{\boldsymbol{\tau}\sim f}[\mathrm{FP}_j(\boldsymbol{\tau},\boldsymbol{\theta})] = \sum_{i=1}^n{\mathbf{1}_{\{\boldsymbol{y}_i\neq \boldsymbol{e}_j\}}	\mathbb{E}_{\boldsymbol{\tau}\sim f}[\mathbf{1}_{\{\hat{\boldsymbol{y}}_{\boldsymbol{\theta}}(\boldsymbol{x}_i)\in R_j(\boldsymbol{\tau})\}}]},\\
		& 	\mathbb{E}_{\boldsymbol{\tau}\sim f}[\mathrm{FN}_j(\boldsymbol{\tau},\boldsymbol{\theta})] = \sum_{i=1}^n{\mathbf{1}_{\{\boldsymbol{y}_i= \boldsymbol{e}_j\}}	\mathbb{E}_{\boldsymbol{\tau}\sim f}[\mathbf{1}_{\{\hat{\boldsymbol{y}}_{\boldsymbol{\theta}}(\boldsymbol{x}_i)\notin R_j(\boldsymbol{\tau})\}}]}.
	\end{split}
\end{equation*}
Then, we are ready to define a MultiSOL.
\begin{definition}
    Let $s$ be a classification score and $f$ the pdf chosen for $\boldsymbol{\tau}$. Moreover, 
    consider the set of predictions $\{\hat{\boldsymbol{y}}_{\boldsymbol{\theta}}(\boldsymbol{x}_i)\}_{i:n}=\{\hat{\boldsymbol{y}}_{\boldsymbol{\theta}}(\boldsymbol{x}_1),\dots,\hat{\boldsymbol{y}}_{\boldsymbol{\theta}}(\boldsymbol{x}_n)\}$ and corresponding labels $\{\boldsymbol{y}_i\}_{i:n}=\{\boldsymbol{y}_1,\dots,\boldsymbol{y}_n\}$. Thus, we can define the MultiSOL
\begin{equation}\label{eq:sol}
	\ell_{s}\big(\{\hat{\boldsymbol{y}}_{\boldsymbol{\theta}}(\boldsymbol{x}_i)\}_{i:n},\{\boldsymbol{y}_i\}_{i:n}\big)= - \frac{1}{m}\sum_{j=1}^m s\big(\mathbb{E}_{\boldsymbol{\tau}\sim f}[\mathrm{CM}_j(\boldsymbol{\tau},\boldsymbol{\theta})]\big).
\end{equation}
\end{definition}
We have the following.
\begin{proposition}
    MultiSOLs satisfy the property:
    \begin{equation}	\min_{\boldsymbol{\theta}}\ell_{s}\big(\{\hat{\boldsymbol{y}}_{\boldsymbol{\theta}}(\boldsymbol{x}_i)\}_{i:n},\{\boldsymbol{y}_i\}_{i:n}\big)\approx \frac{1}{m} \max_{\boldsymbol{\theta}} \sum_{j=1}^m \mathbb{E}_{\boldsymbol{\tau}\sim f}[s(\mathrm{CM}_j(\boldsymbol{\tau},\boldsymbol{\theta}))],
\end{equation}
where equality is achieved if the score is linear with respect to the entries of the confusion matrices. 
\end{proposition}
\textit{Proof.} The proof is a direct consequence of \cite[Theorem 1]{marchetti2025comprehensive}.

In Section \ref{sec:results}, we will test MultiSOLs and see to what extent they concretely retain the same good properties of the original binary SOLs. Before that, we will discuss the actual computation of MultiSOLs.

\subsection{Computing the MultiSOL}\label{sec:monte}

To make the score-oriented loss $\ell_{s}$ in \eqref{eq:sol} operative, we need to calculate $\mathbb{E}_{\boldsymbol{\tau}\sim f}[\mathrm{CM}_j(\boldsymbol{\tau},\boldsymbol{\theta})]$ for the chosen pdf $f$. In the one-dimensional setting corresponding to the binary case, analogous calculations can be effectively performed by taking into account the cdf $F$. However, in the multidimensional setting, closed forms for cdfs are lacking even for the most common probability densities, and thus they are very often estimated numerically. For our purposes, we then consider the Monte Carlo approach, as we detail in what follows.

From the expected confusion matrix, consider
\begin{equation*}
	\mathbb{E}_{\boldsymbol{\tau}\sim f}[\mathbf{1}_{\{\hat{\boldsymbol{y}}_{\boldsymbol{\theta}}(\boldsymbol{x}_i)\in R_j(\boldsymbol{\tau})\}}]=\mathbb{P}_{\boldsymbol{\tau}}(\{\hat{\boldsymbol{y}}_{\boldsymbol{\theta}}(\boldsymbol{x}_i)\in R_j(\boldsymbol{\tau})\})=\int_{\{\hat{\boldsymbol{y}}_{\boldsymbol{\theta}}(\boldsymbol{x}_i)\in R_j(\boldsymbol{\tau})\}}{f(\boldsymbol{\xi})\mathrm{d}\boldsymbol{\xi}}.
\end{equation*}
Letting $N\in\mathbb{N}$, we sample according to the pdf $f$ the threshold values $\boldsymbol{\tau}_1,\dots,\boldsymbol{\tau}_N$ and compute the Monte Carlo approximation
\begin{equation*}
	\mathbb{E}_{\boldsymbol{\tau}\sim f}[\mathbf{1}_{\{\hat{\boldsymbol{y}}_{\boldsymbol{\theta}}(\boldsymbol{x}_i)\in R_j(\boldsymbol{\tau})\}}]\approx\bar{\mathbb{E}}_{\boldsymbol{\tau}}[\mathbf{1}_{\{\hat{\boldsymbol{y}}_{\boldsymbol{\theta}}(\boldsymbol{x}_i)\in R_j(\boldsymbol{\tau})\}}]=\frac{1}{N}\sum_{r=1}^N{\mathbf{1}_{\{\hat{\boldsymbol{y}}_{\boldsymbol{\theta}}(\boldsymbol{x}_i)\in R_j(\boldsymbol{\tau}_r)\}}}.
\end{equation*}
In order to assess the effectiveness on this approximation method, we can take advantage of a classical approach based on the well-known Hoeffding's inequality, that is
\begin{equation}\label{eq:hoeffding}
	\mathbb{P}(|\bar{\mathbb{E}}_{\boldsymbol{\tau}\sim f}[\mathbf{1}_{\{\hat{\boldsymbol{y}}_{\boldsymbol{\theta}}(\boldsymbol{x}_i)\in R_j(\boldsymbol{\tau})\}}]-\mathbb{E}_{\boldsymbol{\tau}\sim f}[\mathbf{1}_{\{\hat{\boldsymbol{y}}_{\boldsymbol{\theta}}(\boldsymbol{x}_i)\in R_j(\boldsymbol{\tau})\}}]|\ge\varepsilon)\le 2e^{-2N\varepsilon^2}
\end{equation}
for $\varepsilon\ge 0$. Indeed, suppose that we wish to guarantee that the absolute value in \eqref{eq:hoeffding} is less than $\varepsilon$ with probability $1-\delta$, $\delta\in(0,1)$. This is equivalent to 
\begin{equation*}
	\mathbb{P}(|\bar{\mathbb{E}}_{\boldsymbol{\tau}\sim f}[\mathbf{1}_{\{\hat{\boldsymbol{y}}_{\boldsymbol{\theta}}(\boldsymbol{x}_i)\in R_j(\boldsymbol{\tau})\}}]-\mathbb{E}_{\boldsymbol{\tau}\sim f}[\mathbf{1}_{\{\hat{\boldsymbol{y}}_{\boldsymbol{\theta}}(\boldsymbol{x}_i)\in R_j(\boldsymbol{\tau})\}}]|\ge\varepsilon)\le \delta.
\end{equation*}
Using \eqref{eq:hoeffding}, this is implied by the stronger
\begin{equation*}
	2e^{-2N\varepsilon^2}\le \delta,
\end{equation*}
which provides us with a concrete rule for choosing the number of samples $N$, that is
\begin{equation*}
	N\ge \frac{\log(2/\delta)}{2\varepsilon^2}.
\end{equation*}
We observe that this Monte Carlo approximation comes with indicator functions, which we want to avoid to obtain a differentiable well-defined loss. We recall that the output $\hat{\boldsymbol{y}}$ belongs to $R_j(\boldsymbol{\tau}_r)$ if for every $k\neq j$ we have $\hat{y}^j-\hat{y}^k>\tau_r^j-\tau^k_r$ (see \eqref{eq:natural}). In place of this \lq\lq True/False\rq\rq$\;$condition, for each $k\neq j$ we compute
\begin{equation*}
    t^{r}_\lambda(k)=\sigma(\lambda(\hat{y}^j-\hat{y}^k-\tau_r^j+\tau^k_r)),
\end{equation*}
where $\lambda>0$ is a real parameter that rules the steepness of the sigmoid $\sigma$. Consequently, we consider the approximation
\begin{equation}\label{eq:lambda}
    \mathbf{1}_{\{\hat{\boldsymbol{y}}_{\boldsymbol{\theta}}(\boldsymbol{x}_i)\in R_j(\boldsymbol{\tau}_r)\}}\approx \prod_{k\neq j}t^{r}_\lambda(k),
\end{equation}
and thus
\begin{equation*}
    \bar{\mathbb{E}}_{\boldsymbol{\tau}\sim f}[\mathbf{1}_{\{\hat{\boldsymbol{y}}_{\boldsymbol{\theta}}(\boldsymbol{x}_i)\in R_j(\boldsymbol{\tau})\}}]\approx \hat{\mathbb{E}}_{\boldsymbol{\tau}\sim f}[\mathbf{1}_{\{\hat{\boldsymbol{y}}_{\boldsymbol{\theta}}(\boldsymbol{x}_i)\in R_j(\boldsymbol{\tau})\}}]=\frac{1}{N}\sum_{r=1}^N{\prod_{k\neq j}t^{r}_\lambda(k)}.
\end{equation*}
Once $	\hat{\mathbb{E}}_{\boldsymbol{\tau}\sim f}[\mathbf{1}_{\{\hat{\boldsymbol{y}}_{\boldsymbol{\theta}}(\boldsymbol{x}_i)\in R_j(\boldsymbol{\tau})\}}]$ and thus
\begin{equation*}
	\hat{\mathbb{E}}_{\boldsymbol{\tau}\sim f}[\mathrm{CM}_j(\boldsymbol{\tau},\boldsymbol{\theta})] = 
	\begin{pmatrix}
		\hat{\mathbb{E}}_{\boldsymbol{\tau}\sim f}[\mathrm{TN}_j(\boldsymbol{\tau},\boldsymbol{\theta})] & 	\hat{\mathbb{E}}_{\boldsymbol{\tau}\sim f}[\mathrm{FP}_j(\boldsymbol{\tau},\boldsymbol{\theta})] \\
		\hat{\mathbb{E}}_{\boldsymbol{\tau}\sim f}[\mathrm{FN}_j(\boldsymbol{\tau},\boldsymbol{\theta})] & 	\hat{\mathbb{E}}_{\boldsymbol{\tau}\sim f}[\mathrm{TP}_j(\boldsymbol{\tau},\boldsymbol{\theta})]
	\end{pmatrix},
\end{equation*}
is computed for each $j=1,\dots,m$, we can build the approximate multiclass SOL
\begin{equation*}
	\hat{\ell}_{s}\big(\{\hat{\boldsymbol{y}}_{\boldsymbol{\theta}}(\boldsymbol{x}_i)\}_{i:n},\{\boldsymbol{y}_i\}_{i:n}\big)= - \frac{1}{m}\sum_{j=1}^m s\big(\hat{\mathbb{E}}_{\boldsymbol{\tau}\sim f}[\mathrm{CM}_j(\boldsymbol{\tau},\boldsymbol{\theta})]\big).
\end{equation*}
\begin{remark}
	The Monte Carlo approach is known to be robust with respect to the curse of dimensions, which in our setting might take place when a large number of classes is involved. Moreover, once the pdf is chosen, the sampling can be carried out at the beginning of the training, and there is no need of resampling threshold values at each iteration of the optimization algorithm.
\end{remark}

\section{Results}\label{sec:results}

The experiments in this paper can be run on consumer-grade hardware. In our case, they were conducted on a desktop PC equipped with 64 GB of RAM, a 16-core AMD Ryzen 5950X CPU, and an NVIDIA RTX 3070 GPU with 8 GB of VRAM. The MultiSOL Python package can be found at 
\begin{equation*}
\texttt{https://github.com/edoardolegnaro/ScoreOrientedLosses}.
\end{equation*}

As far as the \textit{a priori} pdf for the threshold $\boldsymbol{\tau}$ on the $(m-1)$-dimensional simplex is concerned, in our tests we experiment with Dirichlet distributions $f=\mathrm{Dir}(\boldsymbol{\alpha})$ \cite{kotz2019continuous} of parameters $\boldsymbol{\alpha}=(\alpha_1,\dots,\alpha_m)$ with $\alpha_i\equiv \alpha\ge 1$ for each $i=1,\dots,m$. Being these pdfs centered in the barycenter of the simplex, we reasonably use the argmax rule to evaluate the classification scores, which is equivalent to set $\boldsymbol{\tau}=(1/m,\dots,1/m)$ in the threshold-based approach, as already remarked. Then, the proposed loss functions are calculated through a differentiable approximation of a one-vs-rest confusion matrix using $N=1024$ Dirichlet samples, according to the analysis in Subsection \ref{sec:monte}.

This Results section is structured as follows, while the discussion of the obtained results is included in Section \ref{sec:discussion}.
\begin{itemize}
    \item 
    In Subsection \ref{sec:solstizi}, we carry out some experiments to analyze the impact on MultiSOL performance of parameters $\alpha$ in the pdf and technical $\lambda$ from Subsection \ref{sec:monte}.
    \item 
    In Subsection \ref{sec:test_pdf}, we move our focus to an important property of MultiSOLs, that is, the influence of the chosen \textit{a priori} pdf on the resulting \textit{a posteriori} distribution of \textit{good} threshold values, meaning that they provide a related large score value.
    \item 
    In Subsection \ref{sec:scores}, we analyze to what extent a MultiSOL built on a chosen classification score leads to its concrete optimization in the resulting performance, compared to other scores. This is the score-oriented nature of the proposed family of losses.
    \item 
    In Subsection \ref{sec:test_sol}, the performance of the MultiSOL is tested in comparison with other state-of-the-art losses, to assess its competitiveness.
\end{itemize}
In this section, we will experiment with various multiclass datasets, which are reported and briefly described in Table \ref{tab:dataset-summary} in Appendix \ref{app:datasets}. These datasets span a variety of scenarios with different levels of difficulty and class imbalance, providing a comprehensive suite of case studies.

\subsection{Ablation studies of relevant MultiSOL parameters}\label{sec:solstizi}
As network model, we consider a Multi-Layer Perceptron (MLP) trained on the MNIST dataset. This MLP is a feedforward neural network that takes flattened 28×28 images as input. It has two hidden layers with Rectified Linear Unit (ReLU) activations (128 neurons, then 64 neurons), followed by a softmax output layer with a neuron count matching the number of classes. We choose accuracy as baseline score for the construction of the MultiSOL, which is optimized using Adam. Results are shown in Table \ref{tab:CE_vs_SOL_DirPar} for different values of the Dirichlet Parameter $\alpha$ and in Table \ref{tab:CE_vs_SOL_lambda} for different choices of $\lambda$. In both tables, the macro F1 score and accuracy on the test set are reported as the parameters of the MultiSOL function are varied, with comparison to the cross-entropy loss provided as baseline metrics.  

\begin{table}[htbp] 
\centering
\caption{Comparison of scores for an MLP trained on MNIST with CE and MultiSOL losses with varying Dirichlet parameter $\alpha$. We fix $\lambda=20$.}
\label{tab:CE_vs_SOL_DirPar}

\begin{tabular}{ccc|cc}
\toprule
\multicolumn{3}{c|}{\textbf{MultiSOL}} & \multicolumn{2}{c}{\textbf{CE}} \\
\midrule
\textbf{$\alpha$} & \textbf{Macro F1} & \textbf{Acc} 
& \textbf{Macro F1} & \textbf{Acc} \\
\midrule
50   & 0.9777 & 0.9780 & 0.9756 & 0.9758 \\
20   & 0.9794 & 0.9796 & 0.9751 & 0.9753 \\
10   & 0.9768 & 0.9768 & 0.9747 & 0.9749 \\
7.5  & 0.9775 & 0.9777 & 0.9733 & 0.9734 \\
5    & 0.9779 & 0.9781 & 0.9737 & 0.9739 \\
2.5  & 0.9779 & 0.9781 & 0.9755 & 0.9757 \\
1    & 0.9766 & 0.9769 & 0.9741 & 0.9743 \\
\bottomrule
\end{tabular}
\end{table}

\begin{table}[ht]
\centering
\caption{Comparison of scores for an MLP trained on MNIST with CE and MultiSOL losses with varying $\lambda$ parameter. We fix $\alpha=1$.}
\label{tab:CE_vs_SOL_lambda} 

\begin{tabular}{ccc|cc}
\toprule
\multicolumn{3}{c|}{\textbf{MultiSOL}} & \multicolumn{2}{c}{\textbf{CE}} \\
\midrule
\textbf{$\lambda$} & \textbf{Macro F1} & \textbf{Acc} 
& \textbf{Macro F1} & \textbf{Acc} \\
\midrule
1000   & 0.59526 & 0.68570 & 0.91622 & 0.91640 \\
200    & 0.81548 & 0.85540 & 0.91645 & 0.91670 \\
100    & 0.81562 & 0.85540 & 0.91787 & 0.91810 \\
50     & 0.91629 & 0.91640 & 0.91523 & 0.91550 \\
20     & 0.91867 & 0.91890 & 0.91623 & 0.91680 \\
10     & 0.91701 & 0.91710 & 0.91376 & 0.91400 \\
5      & 0.91237 & 0.91270 & 0.91390 & 0.91390 \\
1      & 0.91806 & 0.91850 & 0.91552 & 0.91620 \\
0.5    & 0.91451 & 0.91480 & 0.91568 & 0.91600 \\
0.25   & 0.91269 & 0.91290 & 0.91426 & 0.91460 \\
\bottomrule
\end{tabular}
\end{table}

\subsection{From \textit{priori} to \textit{posteriori}: the influence of the probability density function}\label{sec:test_pdf}

\begin{figure}[htbp]
    \centering
    \includegraphics[width=0.7\linewidth]{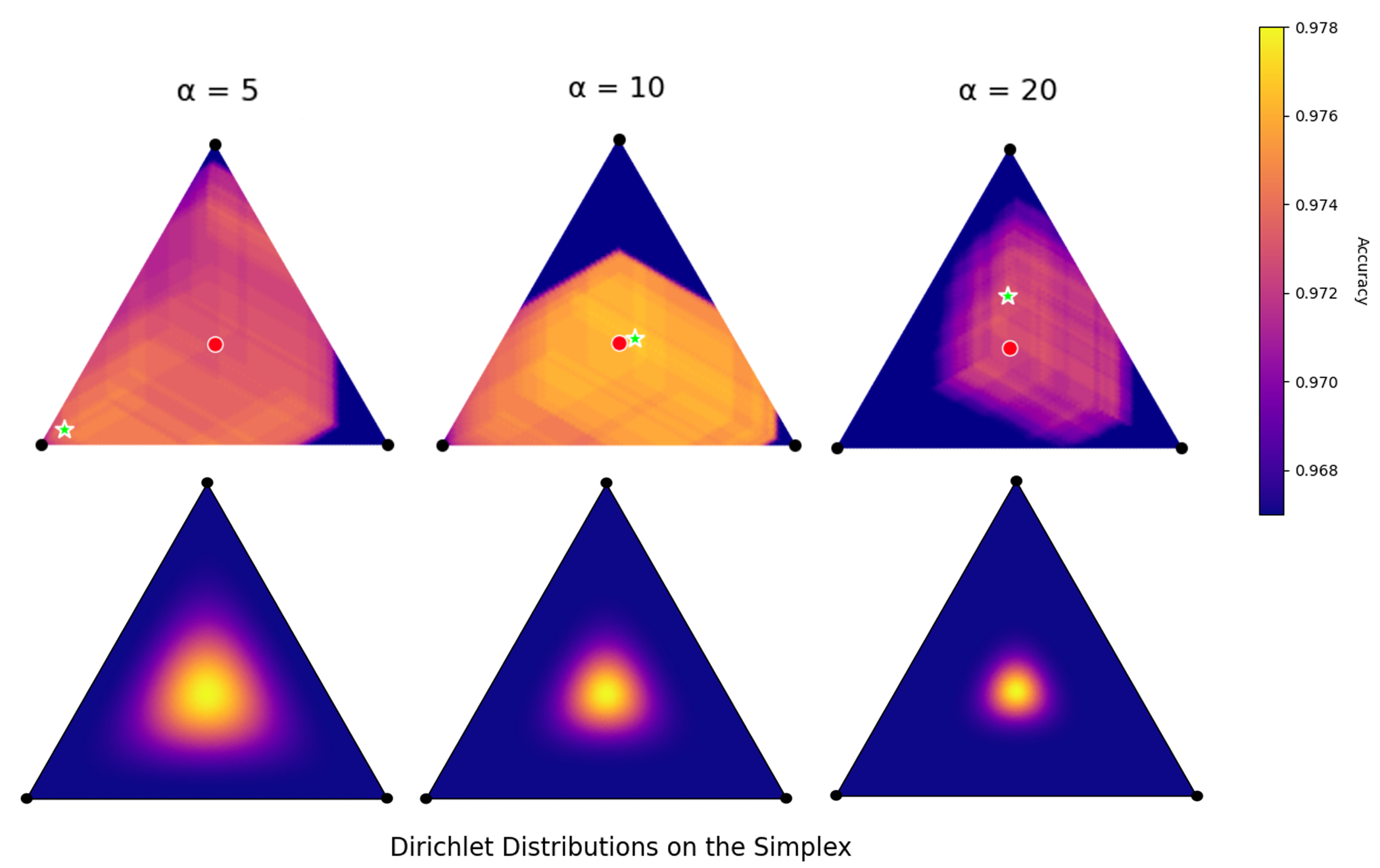}
    \caption{Heatmap of classification accuracies on the simplex (top row) and corresponding a priori Dirichlet distributions (bottom row) for varying parameters $\alpha = 5,10,20$.}
    \label{fig:SolSimplexDirichlet}
\end{figure}

In the considered Dirichlet distributions, we note that $\alpha=1$ leads to a uniform distribution, while the larger $\alpha>1$, the more concentrated pdf is provided around the barycenter of the simplex. This behavior is displayed in Figure \ref{fig:SolSimplexDirichlet} (bottom row) in the $2$-simplex, with $\alpha=5,10,20$. These three \textit{a priori} distributions are considered in training a classifier on the FashionMNIST dataset restricted to the first three classes T-shirt/top, Trouser, and Pullover. The classifier is a ResNet10t with approximately 4.9M parameters, implemented using the timm library \cite{rw2019timm}. Then, in Figure \ref{fig:SolSimplexDirichlet} (top row), we display using a heatmap the classification accuracy resulting from various choices of thresholds in the simplex, which lead to different classification regions (see Figure \ref{fig:simplex_points}).

\subsection{MultiSOL scores comparison}\label{sec:scores}

To evaluate the performance of the introduced MultiSOL when varying the target score, we report below the performance on trained models on different MNIST like datasets.
In particular, we consider MNIST \cite{lecun2002gradient}, FashionMNIST \cite{xiao2017fashion}, and multiclass instances of MedMNIST \cite{medmnistv1}: BloodMNIST \cite{bloodmnist}, DermaMNIST \cite{dermamnist1}, OCTMNIST \cite{octmnist}, OrganAMNIST, OrganCMNIST, OrganSMNIST \cite{xu2019efficient}, PathMNIST \cite{pathmnist} and TissueMNIST \cite{tissuemnist}. Images are $28\times 28$, and no data augmentations are used. We recall that datasets information are summarized in Table \ref{tab:dataset-summary}.

Two architectures are used, depending on the dataset:
\begin{itemize}
    \item \textbf{MNIST and FashionMNIST:} a compact convolutional network with two convolutional blocks (Conv--ReLU--Pool) followed by fully connected layers.
    \item \textbf{MedMNIST datasets:} a lightweight ResNet-style model with four residual stages (64--128--256--512 channels), global average pooling, and a final linear classifier.
\end{itemize}
The architecture for each dataset is fixed across all loss functions to ensure that differences arise solely from the training objective.

For each training, we compare:
\begin{itemize}
    \item \textbf{Cross-entropy (CE):} standard categorical cross-entropy.
    \item \textbf{Weighted CE:} weighted categorical cross-entropy where the class weights are computed from training labels using scikit-learn's balanced strategy.
    \item \textbf{MultiSOL\_score:} proposed loss functions, calculated with $\alpha=1$ and $\lambda=10$. We test MultiSOLs obtained using various underlying scores that are then considered in the classification performance evaluation: accuracy, precision, recall, and F1. Separate models are trained for each MultiSOL variant.
\end{itemize}

All experiments use PyTorch Lightning with Adam (learning rate $10^{-4}$), mixed precision, gradient accumulation for an effective batch size of 128. Each run trains for up to 500 epochs with early stopping (patience 25) based on validation loss. The best checkpoint according to validation loss is evaluated on the test split. All results are averaged over five independent seeds and reported as mean with range (min--max).

{
\tiny
\setlength{\tabcolsep}{2pt}
\renewcommand{\arraystretch}{0.9}
\begin{longtable}{@{\extracolsep{\fill}}lcccc}
\caption{Validation (top) and test (bottom) performance across datasets (mean and range over 5 seeds). Best mean in each column is in bold.}\label{tab:sol-longtable} \\
\toprule
\multicolumn{5}{c}{\textbf{MultiSOL results}} \\
\midrule
\endfirsthead
\toprule
\multicolumn{5}{c}{\textbf{MultiSOL results (cont.)}} \\
\midrule
\endhead
\bottomrule
\endfoot
\bottomrule
\endlastfoot
\multicolumn{5}{c}{\large\textbf{BloodMNIST}} \\
\addlinespace[0.4em]
\multicolumn{5}{l}{\textbf{Validation}} \\
Model & Accuracy & Macro F1 & Macro Precision & Macro Recall \\
\cmidrule(lr){1-5}
CrossEntropy & 0.9378 (0.9298–0.9430) & 0.9293 (0.9204–0.9349) & 0.9309 (0.9188–0.9355) & 0.9289 (0.9246–0.9347) \\
WeightedCE & 0.9204 (0.9073–0.9375) & 0.9289 (0.9176–0.9448) & 0.9269 (0.9208–0.9387) & 0.9289 (0.9176–0.9448) \\
MultiSOL\_acc & \textbf{0.9444 (0.9404–0.9486)} & \textbf{0.9372 (0.9332–0.9423)} & 0.9424 (0.9379–0.9465) & \textbf{0.9335 (0.9270–0.9409)} \\
MultiSOL\_f1 & 0.9328 (0.8834–0.9483) & 0.9154 (0.8166–0.9433) & 0.9136 (0.8038–0.9438) & 0.9193 (0.8374–0.9434) \\
MultiSOL\_prec & 0.9428 (0.9366–0.9486) & 0.9362 (0.9289–0.9417) & \textbf{0.9464 (0.9419–0.9515)} & 0.9287 (0.9168–0.9387) \\
MultiSOL\_rec & 0.4992 (0.4025–0.6621) & 0.3216 (0.2261–0.4843) & 0.2614 (0.1771–0.4149) & 0.4714 (0.3744–0.6173) \\
\addlinespace[0.4em]
\multicolumn{5}{l}{\textbf{Test}} \\
Model & Accuracy & Macro F1 & Macro Precision & Macro Recall \\
\cmidrule(lr){1-5}
CrossEntropy & 0.9378 (0.9298–0.9430) & 0.9293 (0.9204–0.9349) & 0.9309 (0.9188–0.9355) & 0.9289 (0.9246–0.9347) \\
WeightedCE & 0.9204 (0.9073–0.9375) & 0.9289 (0.9173–0.9446) & 0.9269 (0.9208–0.9387) & 0.9289 (0.9176–0.9448) \\
MultiSOL\_acc & \textbf{0.9444 (0.9404–0.9486)} & \textbf{0.9372 (0.9332–0.9423)} & 0.9424 (0.9379–0.9465) & \textbf{0.9335 (0.9270–0.9409)} \\
MultiSOL\_f1 & 0.9328 (0.8834–0.9483) & 0.9154 (0.8166–0.9433) & 0.9136 (0.8038–0.9438) & 0.9193 (0.8374–0.9434) \\
MultiSOL\_prec & 0.9428 (0.9366–0.9486) & 0.9362 (0.9289–0.9417) & \textbf{0.9464 (0.9419–0.9515)} & 0.9287 (0.9168–0.9387) \\
MultiSOL\_rec & 0.4992 (0.4025–0.6621) & 0.3216 (0.2261–0.4843) & 0.2614 (0.1771–0.4149) & 0.4714 (0.3744–0.6173) \\
\addlinespace[1.0em]
\multicolumn{5}{c}{\large\textbf{DermaMNIST}} \\
\addlinespace[0.4em]
\multicolumn{5}{l}{\textbf{Validation}} \\
Model & Accuracy & Macro F1 & Macro Precision & Macro Recall \\
\cmidrule(lr){1-5}
CrossEntropy & 0.7309 (0.7267–0.7327) & 0.4026 (0.3811–0.4112) & 0.5814 (0.4955–0.6183) & 0.3735 (0.3453–0.3905) \\
WeightedCE & 0.3880 (0.3689–0.4143) & \textbf{0.5514 (0.5247–0.5810)} & 0.5478 (0.5309–0.5680) & \textbf{0.5514 (0.5247–0.5810)} \\
MultiSOL\_acc & \textbf{0.7419 (0.7397–0.7441)} & 0.4467 (0.4244–0.4730) & 0.4894 (0.4615–0.5226) & 0.4235 (0.4003–0.4631) \\
MultiSOL\_f1 & 0.7118 (0.6798–0.7337) & 0.4085 (0.3633–0.5251) & 0.4005 (0.3486–0.5278) & 0.4302 (0.3696–0.5429) \\
MultiSOL\_prec & 0.2397 (0.1082–0.4993) & 0.2628 (0.1980–0.4031) & \textbf{0.6462 (0.5609–0.7081)} & 0.3110 (0.2575–0.4186) \\
MultiSOL\_rec & 0.5356 (0.4843–0.6204) & 0.2156 (0.1736–0.2484) & 0.2024 (0.1532–0.2445) & 0.3534 (0.2854–0.4320) \\
\addlinespace[0.4em]
\multicolumn{5}{l}{\textbf{Test}} \\
Model & Accuracy & Macro F1 & Macro Precision & Macro Recall \\
\cmidrule(lr){1-5}
CrossEntropy & 0.7309 (0.7267–0.7327) & 0.4026 (0.3811–0.4112) & 0.5814 (0.4955–0.6183) & 0.3735 (0.3453–0.3905) \\
WeightedCE & 0.3880 (0.3689–0.4143) & \textbf{0.6028 (0.5808–0.6264)} & 0.5478 (0.5309–0.5680) & \textbf{0.5514 (0.5247–0.5810)} \\
MultiSOL\_acc & \textbf{0.7419 (0.7397–0.7441)} & 0.4467 (0.4244–0.4730) & 0.4894 (0.4615–0.5226) & 0.4235 (0.4003–0.4631) \\
MultiSOL\_f1 & 0.7118 (0.6798–0.7337) & 0.4085 (0.3633–0.5251) & 0.4005 (0.3486–0.5278) & 0.4302 (0.3696–0.5429) \\
MultiSOL\_prec & 0.2397 (0.1082–0.4993) & 0.2628 (0.1980–0.4031) & \textbf{0.6462 (0.5609–0.7081)} & 0.3110 (0.2575–0.4186) \\
MultiSOL\_rec & 0.5356 (0.4843–0.6204) & 0.2156 (0.1736–0.2484) & 0.2024 (0.1532–0.2445) & 0.3534 (0.2854–0.4320) \\
\addlinespace[1.0em]
\multicolumn{5}{c}{\large\textbf{FashionMNIST}} \\
\addlinespace[0.4em]
\multicolumn{5}{l}{\textbf{Validation}} \\
Model & Accuracy & Macro F1 & Macro Precision & Macro Recall \\
\cmidrule(lr){1-5}
CrossEntropy & \textbf{0.8942 (0.8942–0.8942)} & \textbf{0.8937 (0.8937–0.8937)} & 0.8938 (0.8938–0.8938) & \textbf{0.8943 (0.8943–0.8943)} \\
MultiSOL\_acc & 0.8851 (0.8806–0.8882) & 0.8839 (0.8785–0.8880) & 0.8846 (0.8808–0.8884) & 0.8847 (0.8799–0.8892) \\
MultiSOL\_f1 & 0.8842 (0.8780–0.8876) & 0.8837 (0.8770–0.8876) & 0.8842 (0.8770–0.8883) & 0.8839 (0.8774–0.8877) \\
MultiSOL\_prec & 0.8840 (0.8795–0.8874) & 0.8831 (0.8780–0.8861) & 0.8842 (0.8803–0.8870) & 0.8838 (0.8788–0.8870) \\
MultiSOL\_rec & 0.5049 (0.4337–0.5578) & 0.5438 (0.4637–0.5884) & \textbf{0.9118 (0.9087–0.9155)} & 0.5058 (0.4360–0.5552) \\
\addlinespace[0.4em]
\multicolumn{5}{l}{\textbf{Test}} \\
Model & Accuracy & Macro F1 & Macro Precision & Macro Recall \\
\cmidrule(lr){1-5}
CrossEntropy & \textbf{0.8859 (0.8859–0.8859)} & \textbf{0.8854 (0.8854–0.8854)} & 0.8854 (0.8854–0.8854) & \textbf{0.8859 (0.8859–0.8859)} \\
MultiSOL\_acc & 0.8733 (0.8687–0.8768) & 0.8722 (0.8669–0.8757) & 0.8728 (0.8682–0.8758) & 0.8733 (0.8687–0.8768) \\
MultiSOL\_f1 & 0.8712 (0.8685–0.8739) & 0.8709 (0.8686–0.8738) & 0.8716 (0.8686–0.8741) & 0.8712 (0.8685–0.8739) \\
MultiSOL\_prec & 0.8724 (0.8680–0.8770) & 0.8717 (0.8689–0.8757) & 0.8727 (0.8703–0.8763) & 0.8724 (0.8680–0.8770) \\
MultiSOL\_rec & 0.5042 (0.4374–0.5553) & 0.5427 (0.4665–0.5889) & \textbf{0.9061 (0.9001–0.9107)} & 0.5042 (0.4374–0.5553) \\
\addlinespace[1.0em]
\multicolumn{5}{c}{\large\textbf{MNIST}} \\
\addlinespace[0.4em]
\multicolumn{5}{l}{\textbf{Validation}} \\
Model & Accuracy & Macro F1 & Macro Precision & Macro Recall \\
\cmidrule(lr){1-5}
CrossEntropy & \textbf{0.9790 (0.9790–0.9790)} & \textbf{0.9788 (0.9788–0.9788)} & \textbf{0.9789 (0.9789–0.9789)} & \textbf{0.9788 (0.9788–0.9788)} \\
MultiSOL\_acc & 0.9777 (0.9741–0.9807) & 0.9776 (0.9739–0.9803) & 0.9777 (0.9742–0.9806) & 0.9775 (0.9738–0.9802) \\
MultiSOL\_f1 & 0.9772 (0.9742–0.9807) & 0.9771 (0.9740–0.9803) & 0.9772 (0.9740–0.9807) & 0.9770 (0.9742–0.9801) \\
MultiSOL\_prec & 0.9776 (0.9748–0.9799) & 0.9774 (0.9746–0.9796) & 0.9775 (0.9746–0.9799) & 0.9775 (0.9747–0.9794) \\
MultiSOL\_rec & 0.9777 (0.9760–0.9794) & 0.9775 (0.9758–0.9791) & 0.9776 (0.9758–0.9792) & 0.9775 (0.9759–0.9792) \\
\addlinespace[0.4em]
\multicolumn{5}{l}{\textbf{Test}} \\
Model & Accuracy & Macro F1 & Macro Precision & Macro Recall \\
\cmidrule(lr){1-5}
CrossEntropy & \textbf{0.9793 (0.9793–0.9793)} & \textbf{0.9792 (0.9792–0.9792)} & \textbf{0.9793 (0.9793–0.9793)} & \textbf{0.9791 (0.9791–0.9791)} \\
MultiSOL\_acc & 0.9769 (0.9736–0.9792) & 0.9767 (0.9733–0.9790) & 0.9769 (0.9738–0.9790) & 0.9766 (0.9730–0.9790) \\
MultiSOL\_f1 & 0.9778 (0.9764–0.9800) & 0.9776 (0.9762–0.9799) & 0.9778 (0.9765–0.9800) & 0.9775 (0.9761–0.9798) \\
MultiSOL\_prec & 0.9779 (0.9733–0.9802) & 0.9778 (0.9731–0.9801) & 0.9779 (0.9733–0.9803) & 0.9778 (0.9731–0.9800) \\
MultiSOL\_rec & 0.9777 (0.9741–0.9799) & 0.9776 (0.9740–0.9798) & 0.9777 (0.9743–0.9798) & 0.9776 (0.9740–0.9799) \\
\addlinespace[1.0em]
\multicolumn{5}{c}{\large\textbf{OCTMNIST}} \\
\addlinespace[0.4em]
\multicolumn{5}{l}{\textbf{Validation}} \\
Model & Accuracy & Macro F1 & Macro Precision & Macro Recall \\
\cmidrule(lr){1-5}
CrossEntropy & 0.7820 (0.7650–0.7980) & 0.7572 (0.7274–0.7777) & 0.8278 (0.8179–0.8391) & 0.7820 (0.7650–0.7980) \\
WeightedCE & 0.8033 (0.7824–0.8283) & 0.8090 (0.7920–0.8320) & 0.8090 (0.7920–0.8320) & 0.8090 (0.7920–0.8320) \\
MultiSOL\_acc & 0.8032 (0.8000–0.8070) & 0.7899 (0.7848–0.7940) & 0.8370 (0.8327–0.8416) & 0.8032 (0.8000–0.8070) \\
MultiSOL\_f1 & 0.8190 (0.8100–0.8320) & 0.8116 (0.7998–0.8238) & 0.8464 (0.8383–0.8539) & 0.8190 (0.8100–0.8320) \\
MultiSOL\_prec & 0.6852 (0.6220–0.7430) & 0.6803 (0.6285–0.7499) & 0.8126 (0.8035–0.8220) & 0.6852 (0.6220–0.7430) \\
MultiSOL\_rec & \textbf{0.8468 (0.8400–0.8530)} & \textbf{0.8434 (0.8370–0.8501)} & \textbf{0.8502 (0.8451–0.8536)} & \textbf{0.8468 (0.8400–0.8530)} \\
\addlinespace[0.4em]
\multicolumn{5}{l}{\textbf{Test}} \\
Model & Accuracy & Macro F1 & Macro Precision & Macro Recall \\
\cmidrule(lr){1-5}
CrossEntropy & 0.7820 (0.7650–0.7980) & 0.7572 (0.7274–0.7777) & 0.8278 (0.8179–0.8391) & 0.7820 (0.7650–0.7980) \\
WeightedCE & 0.8033 (0.7824–0.8283) & 0.8033 (0.7824–0.8283) & 0.8090 (0.7920–0.8320) & 0.8090 (0.7920–0.8320) \\
MultiSOL\_acc & 0.8032 (0.8000–0.8070) & 0.7899 (0.7848–0.7940) & 0.8370 (0.8327–0.8416) & 0.8032 (0.8000–0.8070) \\
MultiSOL\_f1 & 0.8190 (0.8100–0.8320) & 0.8116 (0.7998–0.8238) & 0.8464 (0.8383–0.8539) & 0.8190 (0.8100–0.8320) \\
MultiSOL\_prec & 0.6852 (0.6220–0.7430) & 0.6803 (0.6285–0.7499) & 0.8126 (0.8035–0.8220) & 0.6852 (0.6220–0.7430) \\
MultiSOL\_rec & \textbf{0.8468 (0.8400–0.8530)} & \textbf{0.8434 (0.8370–0.8501)} & \textbf{0.8502 (0.8451–0.8536)} & \textbf{0.8468 (0.8400–0.8530)} \\
\addlinespace[1.0em]
\multicolumn{5}{c}{\large\textbf{OrganAMNIST}} \\
\addlinespace[0.4em]
\multicolumn{5}{l}{\textbf{Validation}} \\
Model & Accuracy & Macro F1 & Macro Precision & Macro Recall \\
\cmidrule(lr){1-5}
CrossEntropy & 0.9204 (0.8926–0.9309) & 0.9165 (0.8883–0.9276) & 0.9264 (0.9035–0.9359) & 0.9102 (0.8811–0.9226) \\
WeightedCE & 0.9133 (0.8870–0.9282) & 0.9169 (0.8916–0.9308) & 0.9130 (0.8885–0.9286) & 0.9169 (0.8916–0.9308) \\
MultiSOL\_acc & 0.9270 (0.9247–0.9301) & 0.9221 (0.9193–0.9268) & 0.9273 (0.9193–0.9327) & 0.9186 (0.9126–0.9248) \\
MultiSOL\_f1 & 0.7911 (0.5529–0.9330) & 0.7878 (0.6192–0.9305) & 0.7887 (0.6205–0.9358) & 0.8077 (0.6702–0.9267) \\
MultiSOL\_prec & \textbf{0.9284 (0.9235–0.9327)} & \textbf{0.9272 (0.9218–0.9318)} & \textbf{0.9334 (0.9261–0.9371)} & \textbf{0.9227 (0.9157–0.9299)} \\
MultiSOL\_rec & 0.2851 (0.2451–0.3824) & 0.1568 (0.1244–0.1897) & 0.1094 (0.0862–0.1379) & 0.3210 (0.2697–0.3577) \\
\addlinespace[0.4em]
\multicolumn{5}{l}{\textbf{Test}} \\
Model & Accuracy & Macro F1 & Macro Precision & Macro Recall \\
\cmidrule(lr){1-5}
CrossEntropy & 0.9204 (0.8926–0.9309) & 0.9165 (0.8883–0.9276) & 0.9264 (0.9035–0.9359) & 0.9102 (0.8811–0.9226) \\
WeightedCE & 0.9133 (0.8870–0.9282) & 0.9167 (0.8909–0.9306) & 0.9130 (0.8885–0.9286) & 0.9169 (0.8916–0.9308) \\
MultiSOL\_acc & 0.9270 (0.9247–0.9301) & 0.9221 (0.9193–0.9268) & 0.9273 (0.9193–0.9327) & 0.9186 (0.9126–0.9248) \\
MultiSOL\_f1 & 0.7911 (0.5529–0.9330) & 0.7878 (0.6192–0.9305) & 0.7887 (0.6205–0.9358) & 0.8077 (0.6702–0.9267) \\
MultiSOL\_prec & \textbf{0.9284 (0.9235–0.9327)} & \textbf{0.9272 (0.9218–0.9318)} & \textbf{0.9334 (0.9261–0.9371)} & \textbf{0.9227 (0.9157–0.9299)} \\
MultiSOL\_rec & 0.2851 (0.2451–0.3824) & 0.1568 (0.1244–0.1897) & 0.1094 (0.0862–0.1379) & 0.3210 (0.2697–0.3577) \\
\addlinespace[1.0em]
\multicolumn{5}{c}{\large\textbf{OrganCMNIST}} \\
\addlinespace[0.4em]
\multicolumn{5}{l}{\textbf{Validation}} \\
Model & Accuracy & Macro F1 & Macro Precision & Macro Recall \\
\cmidrule(lr){1-5}
CrossEntropy & 0.8746 (0.8704–0.8779) & 0.8603 (0.8551–0.8647) & 0.8646 (0.8607–0.8679) & 0.8593 (0.8532–0.8638) \\
WeightedCE & 0.8672 (0.8652–0.8699) & 0.8793 (0.8777–0.8811) & 0.8723 (0.8703–0.8741) & 0.8793 (0.8777–0.8811) \\
MultiSOL\_acc & 0.8986 (0.8944–0.9059) & 0.8857 (0.8795–0.8943) & 0.8876 (0.8818–0.8969) & \textbf{0.8866 (0.8803–0.8937)} \\
MultiSOL\_f1 & \textbf{0.8990 (0.8919–0.9052)} & \textbf{0.8877 (0.8813–0.8939)} & \textbf{0.8917 (0.8846–0.8966)} & 0.8862 (0.8795–0.8943) \\
MultiSOL\_prec & 0.8719 (0.8594–0.8900) & 0.8665 (0.8576–0.8789) & 0.8909 (0.8854–0.8951) & 0.8516 (0.8349–0.8715) \\
MultiSOL\_rec & 0.2954 (0.2125–0.4963) & 0.1863 (0.1284–0.3191) & 0.1400 (0.0910–0.2745) & 0.3483 (0.2695–0.4389) \\
\addlinespace[0.4em]
\multicolumn{5}{l}{\textbf{Test}} \\
Model & Accuracy & Macro F1 & Macro Precision & Macro Recall \\
\cmidrule(lr){1-5}
CrossEntropy & 0.8746 (0.8704–0.8779) & 0.8603 (0.8551–0.8647) & 0.8646 (0.8607–0.8679) & 0.8593 (0.8532–0.8638) \\
WeightedCE & 0.8672 (0.8652–0.8699) & 0.8802 (0.8786–0.8820) & 0.8723 (0.8703–0.8741) & 0.8793 (0.8777–0.8811) \\
MultiSOL\_acc & 0.8986 (0.8944–0.9059) & 0.8857 (0.8795–0.8943) & 0.8876 (0.8818–0.8969) & \textbf{0.8866 (0.8803–0.8937)} \\
MultiSOL\_f1 & \textbf{0.8990 (0.8919–0.9052)} & \textbf{0.8877 (0.8813–0.8939)} & \textbf{0.8917 (0.8846–0.8966)} & 0.8862 (0.8795–0.8943) \\
MultiSOL\_prec & 0.8719 (0.8594–0.8900) & 0.8665 (0.8576–0.8789) & 0.8909 (0.8854–0.8951) & 0.8516 (0.8349–0.8715) \\
MultiSOL\_rec & 0.2954 (0.2125–0.4963) & 0.1863 (0.1284–0.3191) & 0.1400 (0.0910–0.2745) & 0.3483 (0.2695–0.4389) \\
\addlinespace[1.0em]
\multicolumn{5}{c}{\large\textbf{OrganSMNIST}} \\
\addlinespace[0.4em]
\multicolumn{5}{l}{\textbf{Validation}} \\
Model & Accuracy & Macro F1 & Macro Precision & Macro Recall \\
\cmidrule(lr){1-5}
CrossEntropy & 0.7224 (0.7091–0.7476) & 0.6781 (0.6617–0.7072) & 0.7002 (0.6909–0.7174) & 0.6804 (0.6665–0.7044) \\
WeightedCE & 0.6631 (0.6504–0.6973) & 0.7064 (0.6945–0.7289) & 0.6893 (0.6847–0.6975) & 0.7064 (0.6945–0.7289) \\
MultiSOL\_acc & \textbf{0.7666 (0.7589–0.7772)} & \textbf{0.7249 (0.7179–0.7362)} & 0.7357 (0.7287–0.7469) & \textbf{0.7216 (0.7144–0.7375)} \\
MultiSOL\_f1 & 0.7119 (0.4681–0.7755) & 0.6710 (0.4429–0.7374) & 0.6758 (0.4507–0.7469) & 0.6810 (0.4794–0.7328) \\
MultiSOL\_prec & 0.5745 (0.4544–0.7479) & 0.5671 (0.4535–0.7147) & \textbf{0.7901 (0.7475–0.8261)} & 0.5334 (0.4280–0.6986) \\
MultiSOL\_rec & 0.2720 (0.1970–0.4146) & 0.1464 (0.1070–0.2009) & 0.1010 (0.0702–0.1468) & 0.3007 (0.2627–0.3515) \\
\addlinespace[0.4em]
\multicolumn{5}{l}{\textbf{Test}} \\
Model & Accuracy & Macro F1 & Macro Precision & Macro Recall \\
\cmidrule(lr){1-5}
CrossEntropy & 0.7224 (0.7091–0.7476) & 0.6781 (0.6617–0.7072) & 0.7002 (0.6909–0.7174) & 0.6804 (0.6665–0.7044) \\
WeightedCE & 0.6631 (0.6504–0.6973) & 0.6975 (0.6832–0.7273) & 0.6893 (0.6847–0.6975) & 0.7064 (0.6945–0.7289) \\
MultiSOL\_acc & \textbf{0.7666 (0.7589–0.7772)} & \textbf{0.7249 (0.7179–0.7362)} & 0.7357 (0.7287–0.7469) & \textbf{0.7216 (0.7144–0.7375)} \\
MultiSOL\_f1 & 0.7119 (0.4681–0.7755) & 0.6710 (0.4429–0.7374) & 0.6758 (0.4507–0.7469) & 0.6810 (0.4794–0.7328) \\
MultiSOL\_prec & 0.5745 (0.4544–0.7479) & 0.5671 (0.4535–0.7147) & \textbf{0.7901 (0.7475–0.8261)} & 0.5334 (0.4280–0.6986) \\
MultiSOL\_rec & 0.2720 (0.1970–0.4146) & 0.1464 (0.1070–0.2009) & 0.1010 (0.0702–0.1468) & 0.3007 (0.2627–0.3515) \\
\addlinespace[1.0em]
\multicolumn{5}{c}{\large\textbf{PathMNIST}} \\
\addlinespace[0.4em]
\multicolumn{5}{l}{\textbf{Validation}} \\
Model & Accuracy & Macro F1 & Macro Precision & Macro Recall \\
\cmidrule(lr){1-5}
CrossEntropy & 0.8748 (0.8507–0.8981) & 0.8318 (0.8038–0.8576) & 0.8337 (0.8101–0.8555) & 0.8392 (0.8109–0.8645) \\
WeightedCE & 0.7993 (0.7699–0.8341) & 0.8480 (0.8208–0.8819) & 0.8083 (0.7787–0.8413) & 0.8480 (0.8208–0.8819) \\
MultiSOL\_acc & 0.9109 (0.9046–0.9167) & 0.8655 (0.8539–0.8778) & 0.8825 (0.8695–0.8885) & 0.8692 (0.8597–0.8792) \\
MultiSOL\_f1 & \textbf{0.9146 (0.9026–0.9245)} & \textbf{0.8780 (0.8644–0.8939)} & \textbf{0.8829 (0.8680–0.8961)} & \textbf{0.8828 (0.8691–0.9012)} \\
MultiSOL\_prec & 0.5069 (0.3685–0.9191) & 0.5263 (0.3994–0.8847) & 0.8800 (0.8677–0.8927) & 0.5142 (0.3942–0.8864) \\
MultiSOL\_rec & 0.4254 (0.3299–0.5163) & 0.2783 (0.1771–0.4216) & 0.2275 (0.1303–0.3748) & 0.4197 (0.3171–0.5423) \\
\addlinespace[0.4em]
\multicolumn{5}{l}{\textbf{Test}} \\
Model & Accuracy & Macro F1 & Macro Precision & Macro Recall \\
\cmidrule(lr){1-5}
CrossEntropy & 0.8748 (0.8507–0.8981) & 0.8318 (0.8038–0.8576) & 0.8337 (0.8101–0.8555) & 0.8392 (0.8109–0.8645) \\
WeightedCE & 0.7993 (0.7699–0.8341) & 0.8538 (0.8318–0.8810) & 0.8083 (0.7787–0.8413) & 0.8480 (0.8208–0.8819) \\
MultiSOL\_acc & 0.9109 (0.9046–0.9167) & 0.8655 (0.8539–0.8778) & 0.8825 (0.8695–0.8885) & 0.8692 (0.8597–0.8792) \\
MultiSOL\_f1 & \textbf{0.9146 (0.9026–0.9245)} & \textbf{0.8780 (0.8644–0.8939)} & \textbf{0.8829 (0.8680–0.8961)} & \textbf{0.8828 (0.8691–0.9012)} \\
MultiSOL\_prec & 0.5069 (0.3685–0.9191) & 0.5263 (0.3994–0.8847) & 0.8800 (0.8677–0.8927) & 0.5142 (0.3942–0.8864) \\
MultiSOL\_rec & 0.4254 (0.3299–0.5163) & 0.2783 (0.1771–0.4216) & 0.2275 (0.1303–0.3748) & 0.4197 (0.3171–0.5423) \\
\addlinespace[1.0em]
\multicolumn{5}{c}{\large\textbf{TissueMNIST}} \\
\addlinespace[0.4em]
\multicolumn{5}{l}{\textbf{Validation}} \\
Model & Accuracy & Macro F1 & Macro Precision & Macro Recall \\
\cmidrule(lr){1-5}
CrossEntropy & \textbf{0.6690 (0.6659–0.6714)} & 0.5474 (0.5397–0.5523) & 0.6090 (0.6051–0.6152) & 0.5420 (0.5385–0.5460) \\
WeightedCE & 0.5565 (0.5517–0.5652) & \textbf{0.6078 (0.5959–0.6232)} & 0.6071 (0.6045–0.6096) & \textbf{0.6078 (0.5959–0.6232)} \\
MultiSOL\_acc & 0.6677 (0.6659–0.6707) & 0.5598 (0.5498–0.5697) & 0.5889 (0.5802–0.5960) & 0.5574 (0.5454–0.5675) \\
MultiSOL\_f1 & 0.6626 (0.6518–0.6690) & 0.5809 (0.5777–0.5836) & 0.5831 (0.5715–0.5912) & 0.5825 (0.5778–0.5869) \\
MultiSOL\_prec & 0.1582 (0.1468–0.1665) & 0.2360 (0.2240–0.2500) & \textbf{0.7188 (0.7108–0.7310)} & 0.2669 (0.2588–0.2765) \\
MultiSOL\_rec & 0.2656 (0.1299–0.5029) & 0.1215 (0.0683–0.2080) & 0.0878 (0.0411–0.1682) & 0.2690 (0.2304–0.3311) \\
\addlinespace[0.4em]
\multicolumn{5}{l}{\textbf{Test}} \\
Model & Accuracy & Macro F1 & Macro Precision & Macro Recall \\
\cmidrule(lr){1-5}
CrossEntropy & \textbf{0.6690 (0.6659–0.6714)} & 0.5474 (0.5397–0.5523) & 0.6090 (0.6051–0.6152) & 0.5420 (0.5385–0.5460) \\
WeightedCE & 0.5565 (0.5517–0.5652) & \textbf{0.6288 (0.6202–0.6401)} & 0.6071 (0.6045–0.6096) & \textbf{0.6078 (0.5959–0.6232)} \\
MultiSOL\_acc & 0.6677 (0.6659–0.6707) & 0.5598 (0.5498–0.5697) & 0.5889 (0.5802–0.5960) & 0.5574 (0.5454–0.5675) \\
MultiSOL\_f1 & 0.6626 (0.6518–0.6690) & 0.5809 (0.5777–0.5836) & 0.5831 (0.5715–0.5912) & 0.5825 (0.5778–0.5869) \\
MultiSOL\_prec & 0.1582 (0.1468–0.1665) & 0.2360 (0.2240–0.2500) & \textbf{0.7188 (0.7108–0.7310)} & 0.2669 (0.2588–0.2765) \\
MultiSOL\_rec & 0.2656 (0.1299–0.5029) & 0.1215 (0.0683–0.2080) & 0.0878 (0.0411–0.1682) & 0.2690 (0.2304–0.3311) \\
\end{longtable}
}

\normalsize
\subsection{Testing MultiSOL vs SOTA losses}
\label{sec:test_sol}
Here, we compare the proposed MultiSOL with several state-of-the-art and recently proposed loss functions. 
We consider the following losses:
\begin{itemize}
    \item 
    categorical Cross Entropy (CE),
    \item 
    the $\alpha$-divergence loss with GAN regularization (GAN) \cite{pmlr-v235-novello24a},
    \item 
    the squared loss (Square),
    \item 
    the Label-Distribution-Aware Margin loss (LDAM) \cite{cao2019learning}, and
    \item 
    the Influence-Balanced loss (IB) \cite{park2021influence}.
\end{itemize}
All methods share the same training setup. In particular, all experiments are carried out using a ResNet-32 architecture (0.46M parameters) on the CIFAR10 dataset. 
We use the Adam optimizer with an initial learning rate of $10^{-3}$ and scheduled decay over the epochs, together with early stopping based on validation performance.
No further hyperparameter tuning is performed specifically in favor of MultiSOL, which is built on the accuracy score.
For each loss function, we run five independent trainings with different random seeds and report statistics on these runs.

Table~\ref{tab:cifar10_losses} reports the mean test accuracy, standard deviation, minimum and maximum values, and empirical range over the five runs. In addition,  we also monitor the epoch at which early stopping is triggered. 
Table~\ref{tab:cifar10_convergence} reports, for each loss, the average convergence epoch $\pm$ standard deviation and the corresponding range over the five runs.

\begin{table}[htbp]
\centering
\begin{tabular}{lccccc}
\toprule
\textbf{Loss} & \textbf{mean} & \textbf{std} & \textbf{min} & \textbf{max} & \textbf{range} \\
\midrule
CE        & 0.8077 & 0.0072 & 0.7995 & 0.8147 & 0.0152 \\
GAN       & 0.7806 & 0.0027 & 0.7779 & 0.7858 & 0.0079 \\
IB        & 0.7829 & 0.0067 & 0.7706 & 0.7891 & 0.0185 \\
LDAM      & 0.8177 & 0.0037 & 0.8129 & 0.8237 & 0.0108 \\
MultiSOL  & \textbf{0.8205} & 0.0051 & 0.8119 & 0.8244 & 0.0125 \\
Square    & 0.8065 & 0.0075 & 0.7984 & 0.8172 & 0.0188 \\
\bottomrule
\end{tabular}
\caption{Test accuracy on CIFAR10 for different loss functions. 
Values are mean, standard deviation, minimum, maximum and range over $5$ runs with different random seeds.}
\label{tab:cifar10_losses}
\end{table}

\begin{table}[htbp]
\centering
\begin{tabular}{lccc}
\toprule
\textbf{Loss} & \textbf{\# runs} & \textbf{avg. epoch $\pm$ std} & \textbf{range} \\
\midrule
Square   & 5 & $26.2 \pm 3.8$  & 22--32 \\
GAN      & 5 & $29.8 \pm 1.9$  & 28--33 \\
MultiSOL      & 5 & $61.0 \pm 9.8$  & 52--74 \\
CE       & 5 & $30.6 \pm 5.3$  & 27--41 \\
IB       & 5 & $51.4 \pm 13.1$ & 37--74 \\
LDAM     & 5 & $37.2 \pm 7.9$  & 28--48 \\
\bottomrule
\end{tabular}
\caption{Convergence statistics on CIFAR10. 
For each loss we report the number of experiments, the average early-stopping epoch $\pm$ standard deviation, and the observed range.}
\label{tab:cifar10_convergence}
\end{table}

\section{Discussion}\label{sec:discussion}

In Subsection \ref{sec:solstizi}, we assessed the performance of MultiSOL varying the technical parameter $\lambda$ and the Dirichlet \textit{a priori} distribution in the simplex. As reported by Tables \ref{tab:CE_vs_SOL_DirPar} and \ref{tab:CE_vs_SOL_lambda}, the MultiSOL is robust with respect to changes in these parameters values, meaning that its performance can vary, but large observable performance plateaux make the tuning of these parameters non critical.

The tests performed in Subsection \ref{sec:test_pdf} show that the \textit{a priori} pdf chosen for the MultiSOL induces some promising observable effects on the \textit{a posteriori} distribution of \textit{good} threshold values corresponding to large classification scores (see Figure \ref{fig:SolSimplexDirichlet}). This effect is well grounded in the probabilistic features of MultiSOLs, although the concrete optimization of the network does not allow for an even clearer link.

Subsection \ref{sec:scores} is devoted to analyze another remarkable property of MultiSOL, that is, its score-oriented nature. Looking at the numerous tests performed, this effect is evident in many classification scenarios: when focusing on the optimization of a certain score, the corresponding MultiSOL built on that score represents a valuable choice to be taken into account when choosing the loss function of the training design, being competitive or even superior to the classical (weighted) cross entropy. We note that score-oriented optimization is not always fully realized, but this can be mainly related to the optimization process, which is far from perfect in many concrete cases.

Finally, the results in Subsection \ref{sec:test_sol} further certify the competitiveness of the proposed MultiSOL with respect to other state-of-the-art loss functions. Indeed, as reported in Table \ref{tab:cifar10_losses}, we observe that MultiSOL slightly outperforms LDAM, which is the strongest baseline in this comparison, and clearly improves over CE and the remaining methods. Moreover, the empirical variance of MultiSOL is comparable to that of CE and LDAM, indicating stable behavior across different random initializations. In terms of convergence speed assessed through the number of epochs, MultiSOL is slightly slower than CE and LDAM, but its training remains well within a reasonable number of epochs and does not introduce pathological optimization behaviour (see Table \ref{tab:cifar10_convergence}).

\section{Conclusions}\label{sec:conclusions}

In this paper, we proposed a new family of loss functions, i.e., MultiSOLs, to address multiclass classification tasks. The theoretical design of these losses required the usage of a recently introduced threshold-based framework in the multidimensional simplex for the multiclass setting, which allowed the definition of MultiSOLs as a natural generalization of the binary SOLs already studied in the recent literature. The extensive classification presented tests show the robustness and competitiveness of MultiSOLs with respect to other state-of-the-art losses, and highlight their capability in optimizing a chosen classification score of interest. In this direction, further improvements in the minimization process for MultiSOL functions represent an interesting future research line, so that the actual convergence to the theoretical formulation in \eqref{eq:resultone} is better realized, analogously to what obtainable in the binary case (cf. \cite[Section 4.1]{Marchetti22}).



{
\small

\bibliography{references}
}

\bmhead{Statements and Declarations}

E.L. was supported by the HORIZON Europe ARCAFF Project, Grant No. 101082164. All authors acknowledge the Gruppo Nazionale per il Calcolo Scientifico - Istituto Nazionale di Alta Matematica (GNCS - INdAM). The authors have no relevant financial or non-financial interests to disclose. 

\clearpage

\appendix

\section{Datasets}\label{app:datasets}

\begin{figure}
    \centering
    \includegraphics[width=\textwidth]{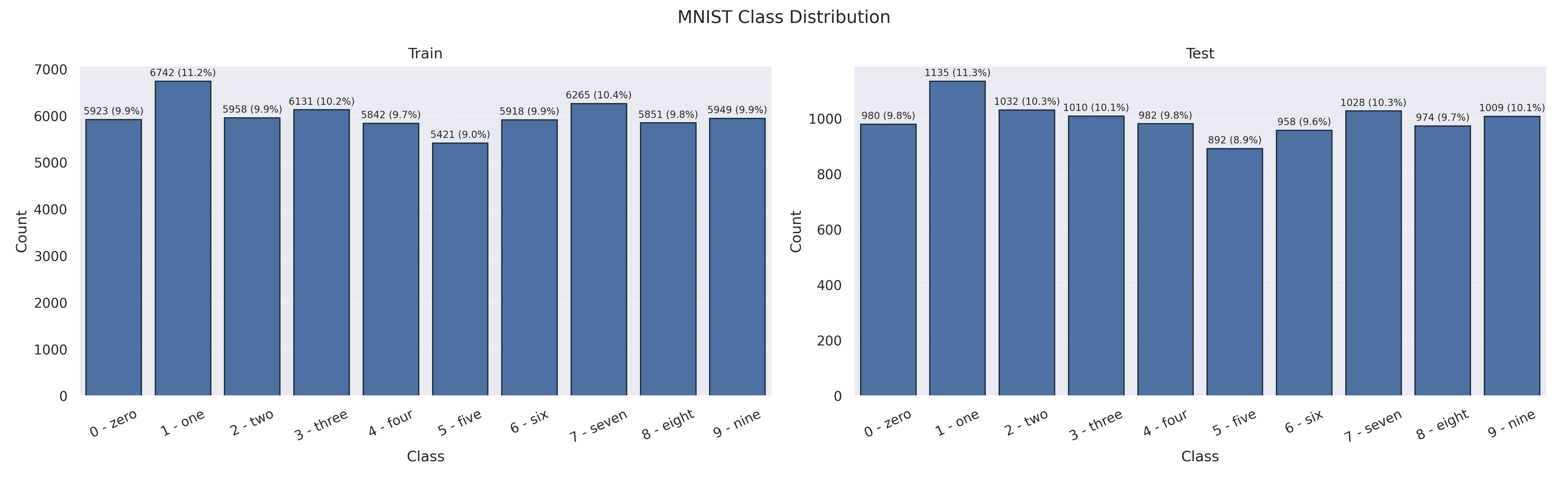}
    \caption{Class distribution of the MNIST dataset for the training and test splits. This shows the expected near-uniform distribution across all ten digits, with only minor fluctuations between classes.}
    \label{fig:mnist_classes}
\end{figure}

\begin{figure}
    \centering
    \includegraphics[width=\textwidth]{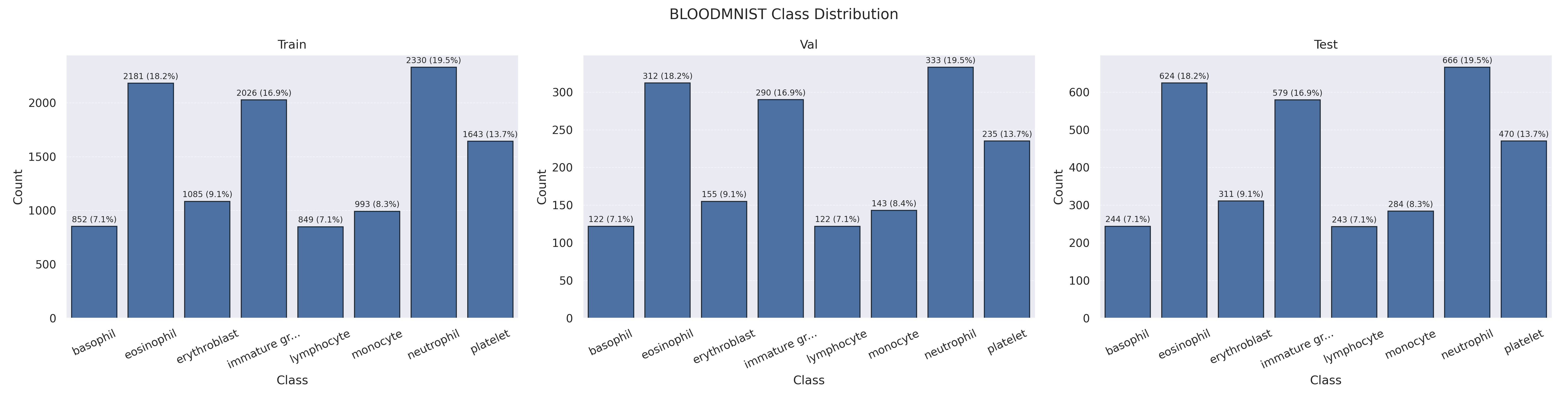}
    \caption{Class distribution of the BloodMNIST dataset across the training, validation, and test splits.  Classes are basophil, eosinophil, erythroblast, immature granulocytes (myelocytes, metamyelocytes and promyelocytes), lymphocyte, monocyte, neutrophil, and platelet. The splits exhibit a consistent moderate imbalance, with neutrophils and eosinophils being most common and basophils and lymphocytes least represented.}
    \label{fig:bloodmnist_classes}
\end{figure}

\begin{figure}
    \centering
    \includegraphics[width=\textwidth]{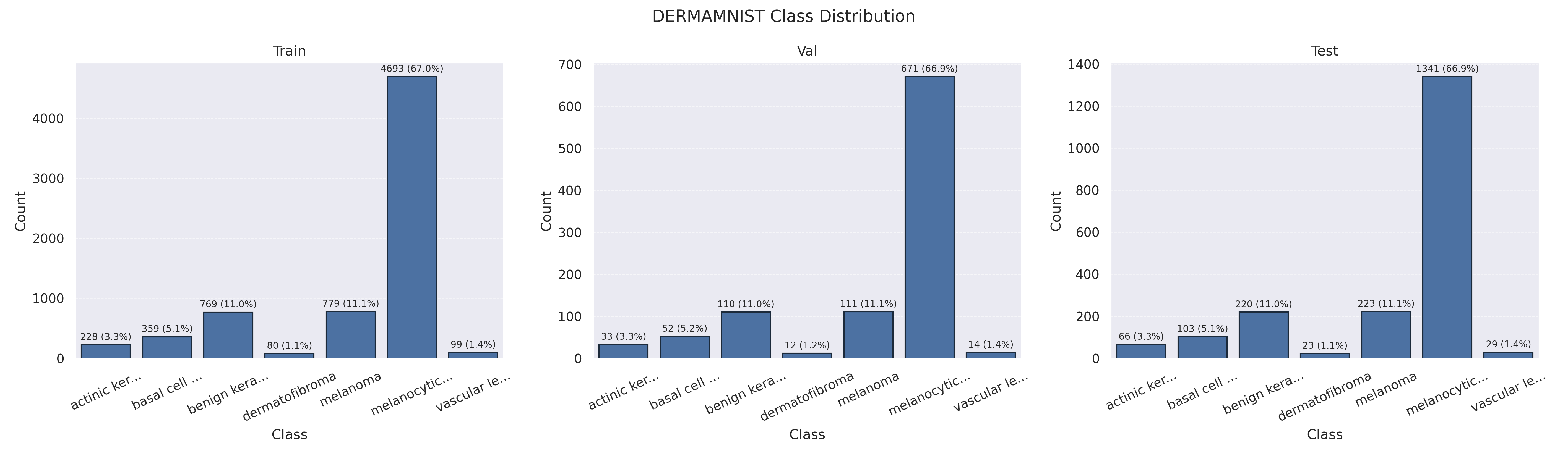}
    \caption{Class distribution of the DermaMNIST dataset across the training, validation, and test splits. The classes are: actinic keratoses and intraepithelial carcinoma, basal cell carcinoma, benign keratosis-like lesions, dermatofibroma, melanoma, melanocytic nevi, and vascular lesions. The dataset is highly imbalanced, dominated by melanocytic nevi (about $67\%$ in all splits), while all other classes occur at much lower frequencies.}
    \label{fig:dermamnist_classes}
\end{figure}

\begin{figure}
    \centering
    \includegraphics[width=\textwidth]{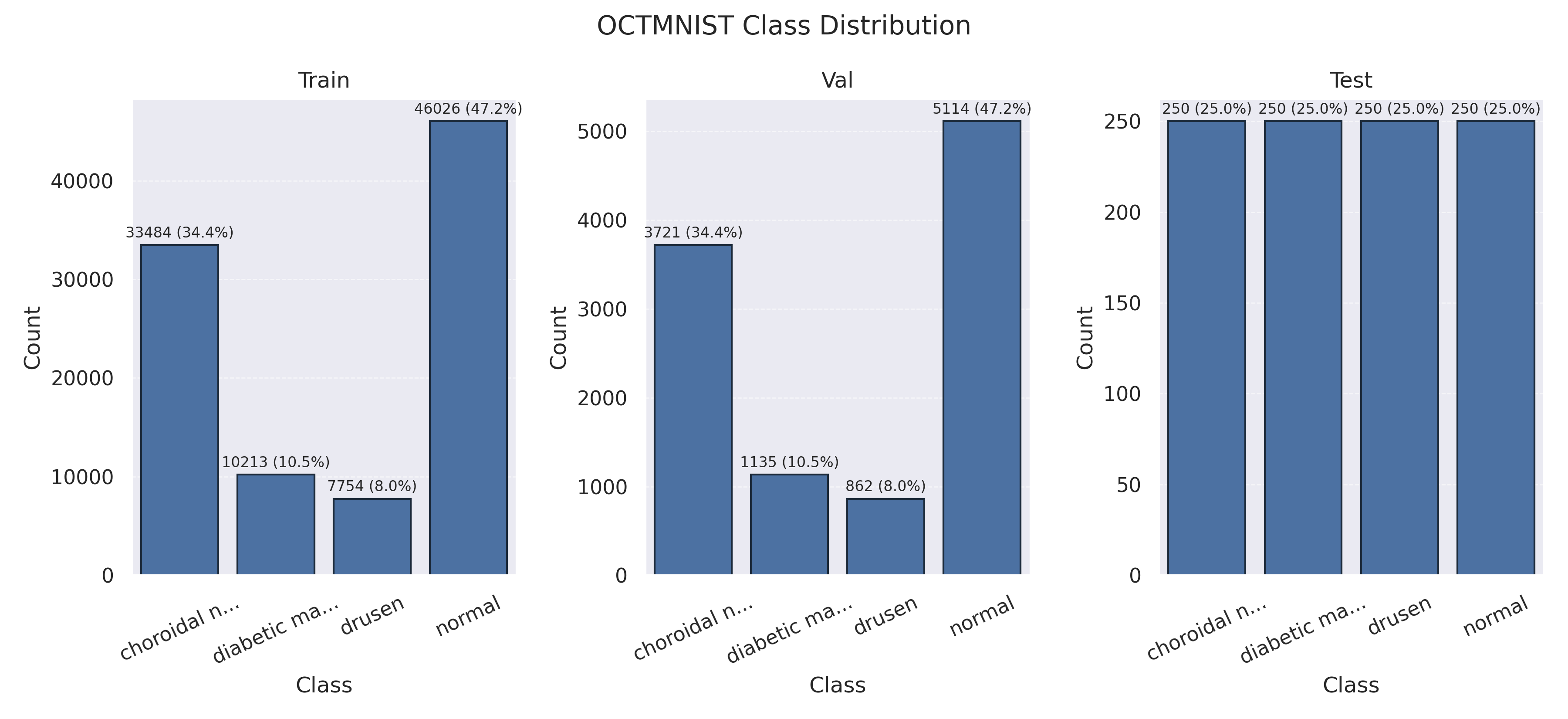}
    \caption{Class distribution of the OCTMNIST dataset for the training, validation, and test splits. The four classes are choroidal neovascularization, diabetic macular edema, drusen, and normal. The training and validation sets are notably imbalanced, with the normal class being the largest group $(\approx 47\%)$ and drusen the smallest $(\approx 8\%)$, while the test set is perfectly balanced across all classes.}
    \label{fig:octmnist_classes}
\end{figure}

\begin{figure}
    \centering
    \includegraphics[width=\textwidth]{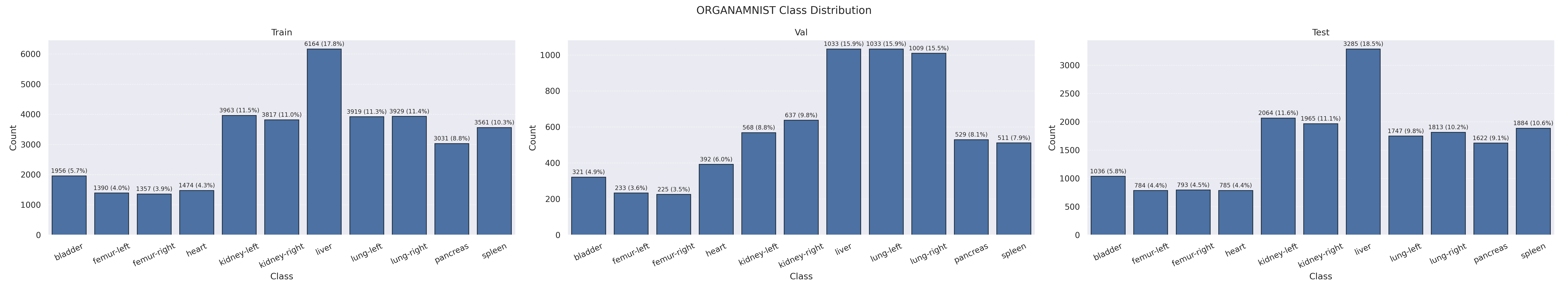}
    \caption{Class distribution of the OrganAMNIST dataset across the training, validation, and test splits. The classes correspond to bladder, femur-left, femur-right, heart, kidney-left, kidney-right, liver, lung-left, lung-right, pancreas, and spleen. Liver is the dominant class throughout all splits $(\approx 16-18\%)$, while femur-right and femur-left constitute the smallest groups $(\approx 3-5\%)$, with the remaining organs distributed in moderate proportions.}
    \label{fig:organamnist_classes}
\end{figure}

\begin{figure}
    \centering
    \includegraphics[width=\textwidth]{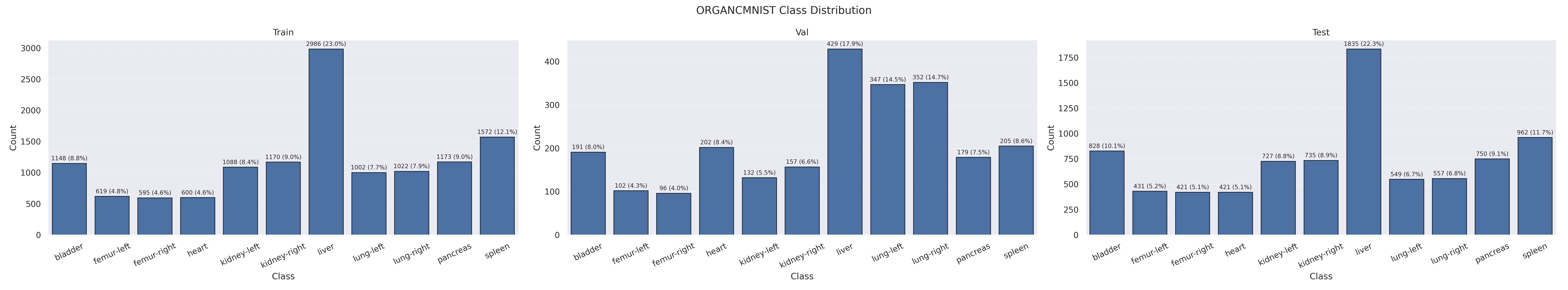}
    \caption{Class distribution of the OrganCMNIST dataset for the training, validation, and test splits. The classes are bladder, femur-left, femur-right, heart, kidney-left, kidney-right, liver, lung-left, lung-right, pancreas, and spleen. Liver is the most frequent class across all splits $(\approx 18-23\%)$, while femur-right and femur-left consistently represent the smallest proportions $(\approx 4-6\%)$. The remaining organs appear in mid-range proportions with similar relative patterns across splits.}
    \label{fig:organcmnist_classes}
\end{figure}

\begin{figure}
    \centering
    \includegraphics[width=\textwidth]{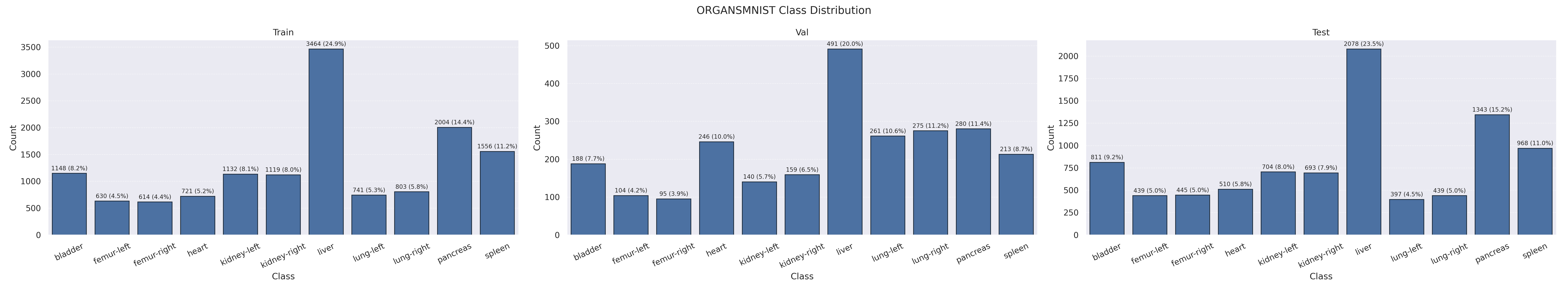}
\caption{Class distribution of the OrganSMNIST dataset across the training, validation, and test splits. The classes are bladder, femur-left, femur-right, heart, kidney-left, kidney-right, liver, lung-left, lung-right, pancreas, and spleen. The liver class is the most frequent across all splits, appearing at approximately $(20\text{--}25\%)$. The smallest classes are femur-left and femur-right, each occurring at roughly $(4\text{--}6\%)$. The remaining organs fall into intermediate frequency ranges, maintaining a consistent imbalance pattern across splits.}
    \label{fig:organsmnist_classes}
\end{figure}

\begin{figure}
    \centering
    \includegraphics[width=\textwidth]{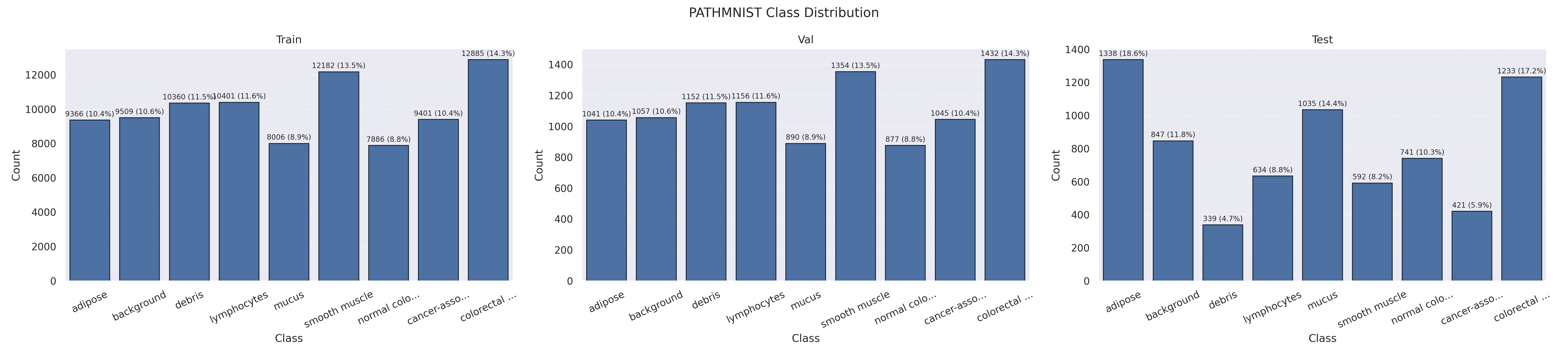}
    \caption{Class distribution of the PathMNIST dataset across the training, validation, and test splits. The classes are adipose, background, debris, lymphocytes, mucus, smooth muscle, normal colon mucosa, cancer-associated stroma, and colorectal adenocarcinoma epithelium. Colorectal adenocarcinoma epithelium is the largest class, occurring at approximately $(14\text{--}17\%)$ across splits. Smallest classes include debris and cancer-associated stroma, each appearing at roughly $(5\text{--}7\%)$. All remaining categories fall into intermediate ranges of approximately $(8\text{--}12\%)$, following a similar imbalance pattern in all splits.}
    \label{fig:pathmnist_classes}
\end{figure}

\begin{figure}
    \centering
    \includegraphics[width=\textwidth]{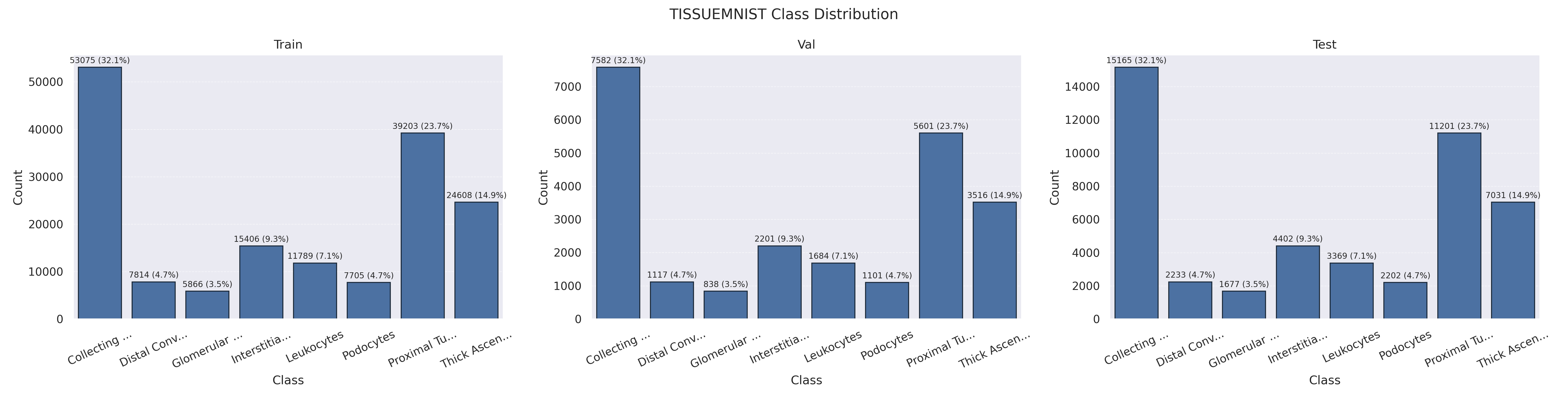}
    \caption{Class distribution of the TissueMNIST dataset across the training, validation, and test splits. The classes are Collecting Duct / Connecting Tubule, Distal Convoluted Tubule, Glomerular endothelial cells, Interstitial endothelial cells, Leukocytes, Podocytes, Proximal Tubule Segments, and Thick Ascending Limb. Collecting Duct / Connecting Tubule is by far the largest class, occurring at approximately $(32\%)$ in all splits. The smallest classes are Glomerular endothelial cells and Distal Convoluted Tubule, each appearing at roughly $(3\text{--}5\%)$. Proximal Tubule Segments consistently form the second-largest category at approximately $(23\text{--}24\%)$, while the remaining classes fall into intermediate ranges of about $(5\text{--}10\%)$.}
    \label{fig:tissue_classes}
\end{figure}

In this appendix we provide a detailed overview of all datasets used in our experimental evaluation in Sections~\ref{sec:scores} and~\ref{sec:test_sol}. We considered standard vision benchmarks (MNIST, FashionMNIST, CIFAR10) and the multiclass medical imaging datasets from the MedMNIST suite (BloodMNIST, DermaMNIST, OCTMNIST, OrganAMNIST, OrganCMNIST, OrganSMNIST, PathMNIST, TissueMNIST), which cover a wide range of numbers of classes, sample sizes, and degrees of class imbalance. Table~\ref{tab:dataset-summary} summarizes, for each dataset, the number of classes and total number of images the total number of images, the train/validation/test split, and qualitative information about the class distribution.
To complement these summaries, Figures~\ref{fig:mnist_classes}–\ref{fig:tissue_classes} provide the full class-frequency histograms for all datasets except FashionMNIST (which is perfectly balanced).
The datasets differ substantially in visual modality and semantic granularity. MNIST consists of $28\times 28$ grayscale images of handwritten digits, while FashionMNIST provides images of clothing items such as shirts, dresses, and footwear, serving as a more visually complex drop-in replacement. CIFAR10 contains natural RGB images of everyday objects and animals across ten broad categories. The MedMNIST datasets span a range of biomedical imaging domains: BloodMNIST includes microscopic blood-cell images; DermaMNIST contains dermatoscopic photographs of skin lesions; OCTMNIST comprises optical coherence tomography scans of retinal tissue; OrganAMNIST, OrganCMNIST, and OrganSMNIST consist of abdominal CT slices depicting different anatomical views (axial, coronal, and sagittal, respectively) of major organs; PathMNIST is based on histopathology tiles of colorectal tissue; and TissueMNIST contains high-resolution microscopy images of kidney cortex cell types.

\begin{table}[h]
    \tiny
    \centering
    \begin{tabular}{lccc}
    \toprule
    Dataset & \# Classes & Total Images & Split \\
    \midrule
    MNIST & 10 & 70{,}000 & 60{,}000 / -- / 10{,}000 \\
    \midrule
    FashionMNIST & 10 & 70{,}000 & 60{,}000 / -- / 10{,}000 \\ 
    \midrule
    BloodMNIST & 8 & 17{,}092 & 11{,}959 / 1{,}712 / 3{,}421 \\
    \midrule
    DermaMNIST & 7 & 10{,}015 & 7{,}007 / 1{,}003 / 2{,}005 \\
    \midrule
    OCTMNIST & 4 & 109{,}309 & 97{,}477 / 10{,}832 / 1{,}000 \\
    \midrule
    OrganAMNIST & 11 & 58{,}830 & 34{,}561 / 6{,}491 / 17{,}778 \\
    \midrule
    OrganCMNIST & 11 & 23{,}583 & 12{,}975 / 2{,}392 / 8{,}216 \\
    \midrule
    OrganSMNIST & 11 & 25{,}211 & 13{,}932 / 2{,}452 / 8{,}827 \\
    \midrule
    PathMNIST & 9 & 107{,}180 & 89{,}996 / 10{,}004 / 7{,}180 \\
    \midrule
    TissueMNIST & 8 & 236{,}386 & 165{,}466 / 23{,}640 / 47{,}280 \\
    \midrule
    CIFAR10 & 10 & 60{,}000 & 50{,}000 / -- / 10{,}000 \\
    \bottomrule
    \end{tabular}
    \caption{Summary of datasets used for the MultiSOL evaluation, including the number of classes, total number of images, and dataset splits.}
    \label{tab:dataset-summary}
\end{table}

\backmatter

\end{document}